\documentclass[sn-mathphys,Numbered]{sn-jnl}

\usepackage{graphicx}%
\usepackage{multirow}%
\usepackage{amsmath,amssymb,amsfonts}%
\usepackage{amsthm}%
\usepackage{mathrsfs}%
\usepackage[title]{appendix}%
\usepackage[table]{xcolor}%
\usepackage{textcomp}%
\usepackage{manyfoot}%
\usepackage{booktabs}%
\usepackage{algorithm}%
\usepackage{algorithmicx}%
\usepackage{algpseudocode}%
\usepackage{listings}%

\newtheorem{theorem}{Theorem}
\newtheorem{proposition}{Proposition}%

\newtheorem{lemma}{Lemma}

\algnewcommand{\Input}[1]{%
  \State \textbf{Input:}
  \hspace*{\algorithmicindent}\parbox[t]{.8\linewidth}{\raggedright #1}
}
\algnewcommand{\Parameters}[1]{%
  \State \textbf{Parameters:}
  \hspace*{\algorithmicindent}\parbox[t]{.8\linewidth}{\raggedright #1}
}
\algnewcommand{\Output}[1]{%
  \State \textbf{Output:}
  \hspace*{\algorithmicindent}\parbox[t]{.8\linewidth}{\raggedright #1}
}
\algnewcommand{\Returns}[1]{%
  \State \textbf{Return:}
  \hspace*{\algorithmicindent}\parbox[t]{.8\linewidth}{\raggedright #1}
}

\begin{document}

\title[Focused PU learning from imbalanced data]{Focused PU learning from imbalanced data}


\author*[1]{\fnm{Elias} \sur{Zavitsanos}}\email{izavits@iit.demokritos.gr}

\author[1]{\fnm{Georgios} \sur{Paliouras}}\email{paliourg@iit.demokritos.gr}

\affil*[1]{\orgdiv{Institute of Informatics and Telecommunications}, \orgname{NCSR ``Demokritos"}, \orgaddress{\street{Patriarhou Gregoriou E and 27 Neapoleos St.}, \city{Agia Paraskevi}, \postcode{15310}, \state{Attica}, \country{Greece}}}

\abstract{We propose a new method of learning from positive and unlabeled (PU) examples in highly imbalanced datasets. Many real-world problems, such as disease gene identification, targeted marketing, fraud detection, and recommender systems, are hard to address with machine learning methods, due to limited labeled data. Often, training data comprises positive and unlabeled instances, the latter typically being dominated by negative, but including also several positive instances. While PU learning is well-studied, few methods address imbalanced settings or hard-to-detect positive examples that resemble negative ones. Our approach uses a focused empirical risk estimator, incorporating both positive and unlabeled examples to train binary classifiers. Empirical evaluations demonstrate state-of-the-art performance on imbalanced datasets under two labeling mechanisms - selecting positives completely at random (SCAR) and selecting at random (SAR). Beyond these controlled experiments, we demonstrate the value of the proposed method in the real-world application of financial misstatement detection.}




\maketitle

\section{Introduction}\label{sec:introduction}
In machine learning, classification tasks involve identifying instances that belong to one or more distinct classes. Binary classification, in particular, aims at a classifier that separates positive and negative instances, also known as positive-negative (PN) classification. This task is among the most widely studied problems in machine learning. A PN machine learning algorithm typically has access to a fully labeled training dataset that consists of representative positive and negative examples. However, there are cases where the training data contain labeled examples of only the positive class, accompanied by a large pool of unlabeled examples. This is known as the positive-unlabeled (PU) learning setting. The assumption is that unlabeled data contain examples from both the positive and the negative class, and the task is the same as binary classification; namely, the algorithm needs to learn to separate positive from negative examples. This setting poses significant hurdles for traditional supervised learning algorithms, which rely heavily on labeled data to establish a clear distinction between classes.

PU learning has gained much attention in various practical scenarios since positive and unlabeled data arise naturally for several reasons \cite{Frenay2014, Zhou2017, Bekker2020}. For instance, in medical diagnosis, it may be impractical to label all patients as either healthy or diseased \cite{Claesen2015, Claesen2015b}. Moreover, in fraud detection, the labels are usually sparse, since the corresponding authority may delay the investigation of all possible cases, due to time and resource limitations \cite{Stripling2018, Bao2020, Bertomeu2021, izavits2021}. The above are illustrative examples where PU learning is required. However, many more real-world applications require learning from PU data, such as under-reporting (e.g., survey data), recommender systems, disease gene identification, targeted advertising, and more \cite{Galarrage2015, Vasighizaker2018, Zupanc2018, Mignone2020a, Mignone2020b}.

Although PU learning has been widely studied in the literature \cite{Bekker2020}, most of the work assumes that the labeled examples are selected completely at random from the positive examples and that the underlying distribution of positive and negative examples is mostly balanced. Such assumptions simplify the problem and allow PU learning to be addressed as a binary classification task with slight modifications. However, the simplifying assumptions do not hold in many practical applications. In particular, class imbalance naturally arises for several reasons, resulting in a biased subset of labeled examples that are not selected completely at random. Class imbalance can be an intrinsic property of the data or result from limitations to obtain labeled data, such as cost, privacy, and significant human effort. There is evidence that if both classes are well represented and come from non-overlapping distributions, it is possible to obtain good results regardless of class disproportion \cite{Japkowicz2000, Krawczyk2016}. If, on the other hand, the positive class is under-represented and overlaps with the negative one, class imbalance becomes more of a challenge in the PU learning setting.

This paper proposes a new PU learning method, relaxing the traditional assumptions as much as possible. We derive a new risk estimator that incorporates both the positive and the unlabeled examples in the learning process, focusing on the few positive examples that are hard to separate from the negative ones. This approach elegantly supports end-to-end training of binary classifiers without manipulating the original data. We compare the proposed approach with other state-of-the-art methods using 14 publicly available datasets and we also experiment with the real-world task of financial misstatement detection where PU data naturally arises. The experiments provide evidence that the proposed method achieves state-of-the-art results. 

The rest of the paper is structured as follows. Section \ref{sec:related_work} provides an overview of the background in PU learning and the related work. Section \ref{sec:method} presents the steps for deriving the new risk estimator. In section \ref{sec:experimental_setup}, we describe the experimental setup and the datasets, while section \ref{sec:results_and_discussion} presents the empirical results. In Section \ref{sec:real_world}, we demonstrate the performance of the method in the real-world application of financial misstatement detection, and finally, in Section \ref{sec:conclusion}, we conclude the paper and suggest future research directions.

\section{Background and related work}\label{sec:related_work}
This section provides preliminary information regarding PU learning and its relation to other types of learning. Then, we present an overview of PU learning approaches, based on a commonly used categorization schema.

\subsection{Preliminaries}\label{sec:preliminaries}
Being a specialization of binary classification, PU learning inherits the assumptions usually made to enable machine learning, such as that the samples are identically and independently distributed and constitute a representative sample of the real population. Moreover, being a variant of weak supervision, but a distinct setting from semi-supervision, it may also inherit related assumptions, such as the smoothness assumption, the cluster assumption, and the manifold assumption. The first assumes that examples close to each other in the feature space are more likely to share the same label. The cluster assumption is a specialization of the smoothness assumption and hypothesizes that the data tend to form discrete clusters. Therefore, data points in the same cluster are more likely to share the same label. The manifold assumption states that the data lie on a manifold of a lower dimension than that of the original feature space and enables the application of distance-based and density-based methods defined on that manifold for learning classifiers.

Additional assumptions can be made regarding the observation of the labeled examples and the underlying labeling mechanism. Most PU learning methods are based on the Selected Completely At Random (SCAR) assumption, which assumes that the (positive) labeled examples constitute a randomly selected subset of the pool of positives, and each positive example has the same probability of being selected to be labeled \cite{Elkan2008}. This is a strong assumption and is usually not realistic in practice. A more realistic assumption is the Selected At Random (SAR) \cite{Bekker2019} one. In SAR, we assume that the labeled examples are a biased sample from the positive distribution, where the selection bias depends on the feature vector representation of the data points. SAR is a loose assumption for most PU learning applications and can be specialized in the case of the Probabilistic Gap (PG) assumption, which states that positive examples that resemble negative ones are less likely to be labeled. In this work, we evaluate the proposed method using both the SCAR and SAR assumptions.

\subsection{PU learning}\label{sec:pu_learning}
The majority of related work in PU learning can be categorized into three main approaches: two-step methods, biased learning techniques, and class-prior-based methods \cite{Bekker2020}. 

\underline{\textbf{Two-step methods}}: Methods in this category usually rely on the smoothness and clustering assumptions \cite{Bekker2020}, i.e., they assume that data points of different classes ideally occupy different regions in the feature space. Two-step methods typically identify the most reliable negatives in the unlabeled data at the first step, and then they train an algorithm from the newly labeled data. During this step, additional positive examples may also be identified and labeled \cite{Fung2006}. Then, by usually considering only the labeled examples, a classifier can be trained in a supervised way.

Early attempts in the field used ``spy" techniques \cite{Liu2003} by turning some labeled examples into spies and placing them into the unlabeled set. Then, a simple classifier, usually Naive Bayes (NB) or Support Vector Machine (SVM), is trained by treating the unlabeled examples as negative and assuming that the reliable negatives are those whose posterior probability is lower than that of the spies. Then, the Expectation-Maximization algorithm is used for the final classification. Other similar methods that perform identification of negative examples or even augmentation of the positive set and then supervised classifier training have been presented in \cite{Li2005, Li2007, Yu2007}. In addition, there are methods that iteratively try to learn a classifier from PU data. The first such approach was the 1-DNF method \cite{Yu2002, Yu2003} that aimed to identify positive examples based on ``strong" positive features in the corresponding vector representation. Then an incremental SVM was used to learn the classifier. Improvements of this approach include the work proposed in \cite{Peng2008}, while other variations use an iterative SVM and Rocchio classification to build prototype vectors for the labeled and the unlabeled examples \cite{Li2003, Li2010}. Other two-step methods exploit clustering to identify negatives in the unlabeled set \cite{Vasighizaker2018, Mignone2020a, Mignone2020b}. The idea here is that reliable negative examples are selected from the clusters that are further away from the positive clusters \cite{Chaudhari2012}.

A common limitation of these approaches is their dependence on a sufficient set of labeled examples, in order to perform the first step, i.e., to select a good set of spies, to choose strong positive features, to successfully apply the Rocchio principles \cite{Uden2007}, or to properly form separable clusters in the feature space. This limitation makes it difficult for these techniques to be used in settings with class imbalance, where the labeled set comprises very few positive examples.

To deal with the above limitation, the first step can be viewed as an anomaly detection step, and related techniques can be used to identify reliable negatives or positives from the unlabeled set. An initial approach \cite{Li2007} focused on cases where very few negative examples were expected in the unlabeled data. Unfortunately, that setting does not align well with the common class imbalance situation, where the class of interest is the positive one. A more interesting recent approach proposes a method that relies on a modification of the Isolation Forest outlier detection method to find unreliable negatives and either label them as positive or ignore them and transform the problem into a supervised learning setting \cite{Ortega2023}.

\underline{\textbf{Biased learning}}: Techniques in this category rely on the negativity assumption, meaning that they treat unlabeled data as negative with label noise. In this category, work based on a biased SVM has been proposed to penalize misclassified examples differently, depending on whether they are truly positive or negative \cite{Liu2003b}. A direct extension relies on multiple iterations of the biased SVM, in order for misclassified examples to receive extra penalty \cite{Ke2012}. Another modification assigns a weight to each example in the unlabeled set, on top of the misclassification penalty, indicating the likelihood of this example being negative \cite{Liu2005}. Similarly, the weighted logistic regression used in \cite{Lee2003} favors correct positive classification over correct negative classification by using different weights. 

Bagging techniques have also been proposed in this direction, mainly to deal with the bias towards the negative class and to address the fictitious higher importance that may be given to noisy negative examples. The work in \cite{Mordelet2013}, for instance, uses bagging SVM to learn multiple classifiers trained on labeled examples and subsets of the unlabeled examples, and an extension is presented in \cite{Claesen2015}. Other work that is also based on SVMs incorporates additional regularization terms and favors examples that are close to each other and have the same label \cite{Suykens1999, Ke2018, Ke2018b}.

A parallel line of research develops foundation models for tabular data, such as those used most often in the PU learning literature. Most prominently, TabPFN \cite{hollmann2025tabpfn} is a transformer that performs in-context learning over a labeled support set and approximates the Bayesian posterior-predictive distribution in a single forward pass. Such models target the fully supervised positive–negative setting and expose no notion of unlabeled data. Therefore, applying them to PU problems requires treating the unlabeled pool as negative.

\underline{\textbf{Incorporation of class prior}}: Methods that rely on the class prior to either modify the risk estimators or the prediction of the learned model assume that the class prior is known or it can be estimated from data \cite{Bekker2018, Ramaswamy2016, Plessis2017, Zeiberg2020}. Early work in this direction has proposed duplicating all the unlabeled examples in the training set to count them as partially positive and negative based on the class prior probabilities \cite{Elkan2008}. More recent approaches propose modifications to the learning algorithm, using risk minimizers that estimate the misclassification cost for both the positive and the unlabeled examples. Such modifications of the loss function of the algorithm have led to the direct adaptation of several learning algorithms such as logistic regression \cite{Bekker2018, Plessis2014, Plessis2015} and neural networks \cite{Kiryo2017} for learning with PU data. These methods are considered state-of-the-art solutions to PU learning. However, they typically cannot handle cases with significant class imbalance and the challenges that it poses. One exception to this norm is the work in \cite{Su2021}, where a non-negative risk estimation for PU learning (nnPU) was proposed, building on the work in \cite{Plessis2015} and \cite{Kiryo2017}, and assuming that oversampling of positive examples is helpful in imbalanced scenarios.

The incorporation of the class prior has also been exploited in recent work \cite{PUHellinger} that introduces the PU Hellinger Decision Tree (PUHDT) and the PU Hellinger Random Forest (PUHRF). PUHDT is specifically designed for learning from imbalanced PU datasets under the SCAR assumption, without requiring resampling or misclassification costs. It adapts the traditional Hellinger distance-based splitting criterion to handle PU settings, which provides robustness to class imbalance. PUHRF extends PUHDT, by incorporating it as a base learner in an ensemble framework, using a stratified bootstrap sampling to ensure that all labeled positive examples are included in the training data. The above methods rely on SCAR to estimate the number of positive examples in unlabeled data nodes. Under the SAR or PG assumption, this estimation becomes biased and unreliable, leading to inaccurate splits and decisions in the tree. Therefore, PUHDT and PUHRF are tailored to model PU data under the SCAR assumption only.

The family of the nnPU risk estimators provides elegant PU learning methods that do not rely on data manipulation and preprocessing. Moreover, it offers a way to experiment with appropriate loss functions that can help with class imbalance and deal with hard-to-classify examples. In this work, we adopt this approach to derive a risk estimator, based on imbalanced PU data. The new estimator focuses on the few positive examples that are difficult to distinguish because of the Probabilistic Gap assumption. Such methods provide flexibility because they can be incorporated into neural networks and can be easily extended to semi-supervised settings beyond PU.

\section{Method}\label{sec:method}
\subsection{Background and notation}
The work of du Plessis et al. \cite{Plessis2014, Plessis2015, Plessis2017} and the related theoretical analysis \cite{Niu2016} proposed unbiased risk estimators for PU classification (uPU), by utilizing unlabeled data for risk evaluation. The implication is that label information can be extracted from unlabeled data, without regularization or other restrictive assumptions. Together with the extension to non-negative risk estimators (nnPU) in \cite{Kiryo2017}, and the work in \cite{Su2021}, they revolutionized the field of PU learning. 

The above work has shown that in a PU setting, where negative examples are not available, the corresponding empirical risk can be approximated indirectly. The derivation of the PU risk estimator is briefly described as follows. Let $\boldsymbol{x} \in \mathbb{R}^{d}$ and $y = \{+1, -1\}$, where $d$ is a positive integer, be the input and output variables respectively with probability density $p(\boldsymbol{x},y)$. In a PN classification setting from $\boldsymbol{x}$ to $y$, let us consider two sets of samples, namely the positive $(\mathcal{P})$ and the negative $\mathcal{(N)}$:

\begin{equation*}
\begin{split}
    \mathcal{X}_{P}:=\{\boldsymbol{x}_{i}^{P}\}_{i=1}^{n_P} \overset{\mathrm{iid}}{\sim} p_{P}(\boldsymbol{x}):=p(\boldsymbol{x}|y=+1), \\
     \mathcal{X}_{N}:=\{\boldsymbol{x}_{i}^{N}\}_{i=1}^{n_N} \overset{\mathrm{iid}}{\sim} p_{N}(\boldsymbol{x}):=p(\boldsymbol{x}|y=-1)
    \end{split}
\end{equation*}

Also, $\pi_{P}:=p(y=+1)$ and $\pi_{N}:=p(y=-1)$ are the class prior probabilities for the positive and negative classes such that $\pi_{P}+\pi_{N}=1$.

Let $g: \mathbb{R}^{d} \rightarrow \mathbb{R}$ be a decision function for PN classification. If $\ell:\mathbb{R} \times \{\pm 1\} \rightarrow \mathbb{R}$ is a continuous surrogate loss function, then the partial risks of the classifier $g$ under loss $\ell$ are:

\begin{equation*}
    \begin{split}
        \mathcal{R}_{P}^{+}(g) := \mathbb{E}_{P}[\ell(g(\boldsymbol{x}), +1)] = \mathbb{E}_{P}[\ell(g(\boldsymbol{x}))], \\
        \mathcal{R}_{N}^{-}(g) := \mathbb{E}_{N}[\ell(g(\boldsymbol{x}), -1)] = \mathbb{E}_{N}[\ell(-g(\boldsymbol{x}))]
    \end{split}
\end{equation*}
where $\mathbb{E}_{P}$ and $\mathbb{E}_{N}$ denote the expectations over $p_{P}(\boldsymbol{x})$ and $p_{N}(\boldsymbol{x})$ respectively.

In standard PN classification, where fully labeled training data are available, the goal is to minimize the risk of $g$ with respect to $\ell$ under $p(\boldsymbol{x},y)$, which is defined as:

\begin{equation}
   \mathcal{R}(g):=\mathbb{E}_{(x,y)}[\ell(g(\boldsymbol{x}), y)]
\end{equation}
and can be approximated directly from data by:

\begin{equation}\label{eq:pn_risk}
    \begin{split}
    \mathcal{\tilde{R}}_{PN}(g) = \pi_{P}\mathcal{\tilde{R}}_{P}^{+}(g)+\pi_{N}\mathcal{\tilde{R}}_{N}^{-}(g) \\
    = \pi_{P}\mathcal{\tilde{R}}_{P}^{+}(g)+(1 - \pi_{P})\mathcal{\tilde{R}}_{N}^{-}(g),
    \end{split}
\end{equation}
where:

\begin{equation*}
\begin{split}
    \mathcal{\tilde{R}}_{P}^{+}(g) = \dfrac{1}{n_{P}} \sum_{i=1}^{n_{P}} \ell(g(\boldsymbol{x}_{i}^{P})), \\
    \mathcal{\tilde{R}}_{N}^{-}(g) = \dfrac{1}{n_{N}} \sum_{i=1}^{n_{N}} \ell(-g(\boldsymbol{x}_{i}^{N}))
    \end{split}
\end{equation*}
By minimizing $\mathcal{\tilde{R}}_{PN}(g)$ we obtain the ordinary empirical risk minimizer $\tilde{g}_{PN}$.

On the other hand, in PU learning $\mathcal{X}_{N}$ is not available, thus $\mathcal{R}_{N}^{-}(g)$ cannot be directly estimated. Instead of $\mathcal{X}_{N}$, we have access to an unlabeled set $\mathcal{(U)}$:

\begin{equation*}
    \mathcal{X}_{U}:=\{\boldsymbol{x}_{i}^{U}\}_{i=1}^{n_U} \overset{\mathrm{iid}}{\sim}p(\boldsymbol{x}) :=\pi_{P}\times p_{P}(\boldsymbol{x})+\pi_{N} \times p_{N}(\boldsymbol{x})
\end{equation*}

However, it has been shown \cite{Plessis2015} that we can estimate $\mathcal{R}(g)$ without any bias by replacing the second term of Eq. \ref{eq:pn_risk}. This can be done by exploiting the fact that 
\begin{equation*}
    p(\boldsymbol{x}) = \pi_{P}\times p_{P}(\boldsymbol{x})+(1 - \pi_{P}) \times p_{N}(\boldsymbol{x})
\end{equation*}
and therefore use the following equality:

\begin{equation}
    (1-\pi_{P}) \mathcal{R}_{N}^{-}(g) = \mathcal{R}_{U}^{-}(g) - \pi_{P} \mathcal{R}_{P}^{-}(g)
\end{equation}
where:

\begin{equation*}
\begin{split} 
    \mathcal{R}_{U}^{-}(g):=\mathbb{E}_{U}[\ell(g(x), -1)], \\
    \mathcal{R}_{P}^{-}(g):=\mathbb{E}_{P}[\ell(g(x), -1)]
    \end{split}
\end{equation*}
are the empirical risks of the unlabeled samples and the positive samples in the unlabeled set, respectively. As a result, the classification risk can be approximated by:

\begin{equation}\label{eq:uPU_empirical_risk}
    \mathcal{\tilde{R}}_{PU}(g):=\pi_{P}\mathcal{\tilde{R}}_{P}^{+}(g)-\pi_{P}\mathcal{\tilde{R}}_{P}^{-}(g)+\mathcal{\tilde{R}}_{U}^{-}(g),
\end{equation}
where:

\begin{equation*}
    \begin{split}
        \mathcal{\tilde{R}}_{P}^{-}(g) = \dfrac{1}{n_{P}} \sum_{i=1}^{n_{P}} \ell(-g(\boldsymbol{x}_{i}^{P})), \\
        \mathcal{\tilde{R}}_{U}^{-}(g) = \dfrac{1}{n_{U}} \sum_{i=1}^{n_{U}} \ell(-g(\boldsymbol{x}_{i}^{U}))
    \end{split}
\end{equation*}
By minimizing $\mathcal{\tilde{R}}_{PU}(g)$ we obtain the empirical minimizer $\tilde{g}_{PU}$. The authors note that if a symmetric loss function is chosen, such as the sigmoid loss that was used in \cite{Plessis2015}, the PU risk becomes a cost-sensitive classification of $\mathcal{P}$ and $\mathcal{U}$ data with weight $2\pi_{P}$: $\tilde{\mathcal{R}}_{PU}:=2\pi_{P}\tilde{\mathcal{R}}_{P}^{+}(g)+\tilde{\mathcal{R}}_{U}^{-}(g)-\pi_{P}$.

In practice, it was found that when using neural networks, the second term of Eq. \ref{eq:uPU_empirical_risk} can go much lower than zero, although theoretically it should always be non-negative, because it is used to estimate $(1-\pi_{P})\mathbb{E}_{P}[\ell(g(x), -1)]$. Therefore, the authors in \cite{Kiryo2017, Plessis2017} observed that:

\begin{equation*}
\mathcal{R}_{U}^{-}(g) - \pi_{P} \mathcal{R}_{P}^{-}(g) = (1 - \pi_{P}) \mathcal{R}_{N}^{-}(g) \geq 0
\end{equation*}
and proposed the following non-negative PU risk estimator:

\begin{equation}\label{eq:nnPU}
    \tilde{\mathcal{R}}_{nnPU}(g):=\pi_{P}\tilde{\mathcal{R}}_{P}^{+}(g) + \max \{0,  \tilde{\mathcal{R}}_{U}^{-}(g) - \pi_{P}\tilde{\mathcal{R}}_{P}^{-}(g)\}
\end{equation}

In the following section, we aim to derive a non-negative risk estimator using an appropriate loss function that addresses the challenges of the underlying imbalanced class distribution in the data and the existence of positive examples in the unlabeled set that are hard to separate from negative ones. 

\subsection{Focused risk estimator}\label{sec:iFPU}
In imbalanced data classification, we assume that the positive class is under-represented while the negative class dominates the data. In PU data, this translates to having few labeled positive examples and, at the same time, expecting a few positive examples to be hidden in the unlabeled set. Consequently, the first term of Eq. \ref{eq:nnPU} becomes close to zero and the classification cost is dominated by the unlabeled data. In many real-world domain applications, we also expect to have among the unlabeled examples some positive ones that are hard to classify, since they may be close to negative ones (PG assumption). These two phenomena make the learned model weak in identifying positive examples. Therefore, the proposed risk estimators must be equipped with appropriate loss functions to deal with this situation. In this work, we derive a non-negative risk estimator for PU data that relies on Focal Loss \cite{Lin2020}, which divides samples into hard-to-classify and easy-to-classify. 

The focal loss is based on the cross-entropy loss function, appropriately modified to address the class imbalance problem in classification tasks, by down-weighting the loss for correctly classified examples and up-weighting the loss for misclassified examples. The focal loss is defined in Eq. \ref{eq:FL}, where the exponential term $\gamma$ controls the weight of easy and hard examples and the shape of the loss curve. When the probability for the target class ($p_{t})$ tends to 1 and the true class is the positive one, the example is an easily classifiable example and the contribution to the loss is small. When $p_{t}$ tends to 0 but still the true class is the positive one for the particular example, the contribution to the loss is larger.

\begin{equation}\label{eq:FL}
    \mathcal{FL}(p_{t}) = -(1-p_{t})^{\gamma}log(p_{t})
\end{equation}

To incorporate the focal loss in the empirical risk estimator for PU classification, we rewrite it as a composite loss, considering the ground truth $y$ and the prediction $p$. Let $p_{t}$ of Eq. \ref{eq:FL} be:

\begin{equation}
    p_{t}=
    \begin{cases}
      p, & \text{if           $y=1$}\\
      1-p, & \text{if   $y=-1$}
    \end{cases}
\end{equation}

Then, the focal loss is written as:

\begin{equation}
    \mathcal{FL}=
    \begin{cases}
      -(1-p)^{\gamma}log(p), & \text{if           $y=1$}\\
      -p^{\gamma}log(1-p), & \text{if             $y=-1$}
    \end{cases}
\end{equation}

We can now work from Eq. \ref{eq:uPU_empirical_risk} for the expected risk, by expanding $\mathcal{R}_{P}^{+}(g)$, $\mathcal{R}_{P}^{-}(g)$, and $\mathcal{R}_{U}^{-}(g)$, in order to approximate the true risk. We call the derived risk estimator \textbf{iFPU}:

\begin{equation}
    \mathcal{R}_{iFPU}(g):=\pi_{P}\mathcal{R}_{P}^{+}(g)-\pi_{P}\mathcal{R}_{P}^{-}(g)+\mathcal{R}_{U}^{-}(g)
\end{equation}
and by expanding, we have:

\begin{equation}
    \begin{split}
        \tilde{\mathcal{R}}_{iFPU}(g) = \dfrac{1}{n_{P}}\sum_{i=1}^{n_{P}}- \pi_{P}(1-g(x_{i}))^{\gamma}log(g(x_{i})) \\
        + \dfrac{1}{n_{U}}\sum_{i=1}^{n_{U}}-(g(x_{i}))^{\gamma}log(1-g(x_{i})) \\
        - \dfrac{1}{n_{P}}\sum_{i=1}^{n_{P}}- \pi_{P}(g(x_{i}))^{\gamma}log(1-g(x_{i}))
    \end{split}
\end{equation}

and the corresponding non-negative PU risk is, therefore:

\begin{equation}
    \begin{split}
        \tilde{\mathcal{R}}_{iFPU}(g) = \dfrac{1}{n_{P}}\sum_{i=1}^{n_{P}}- \pi_{P}(1-g(x_{i}))^{\gamma}log(g(x_{i})) \\
        + \max \{0, \dfrac{1}{n_{U}}\sum_{i=1}^{n_{U}}-(g(x_{i}))^{\gamma}log(1-g(x_{i})) \\
        - \dfrac{1}{n_{P}}\sum_{i=1}^{n_{P}}- \pi_{P}(g(x_{i}))^{\gamma}log(1-g(x_{i})) \}
    \end{split}
\end{equation}

In terms of training, stochastic optimization is used. When $\tilde{\mathcal{R}}_{U}^{-}(g) - \pi_{P}\tilde{\mathcal{R}}_{P}^{-}(g)$ becomes smaller than zero for a mini-batch, we perform a step of gradient ascent along $\nabla(\tilde{\mathcal{R}}_{U}^{-}(g) - \pi \tilde{\mathcal{R}}_{P}^{-}(g)$ to make the mini-batch less overfitted. The training process of the non-negative focused risk estimator is performed according to Algorithm \ref{algo1}.

\begin{algorithm}
\caption{Imbalanced Focused PU (iFPU)}\label{algo1}
\begin{algorithmic}[1]
\Input{training data $(\mathcal{X}_{P}, \mathcal{X}_{U})$}
\Parameters{class prior $\pi$, max epochs, learning rate}
\Output{classifier $\tilde{g}_{PU}(x;\theta)$} 
\State Let $\mathcal{A}$ be an external SGD-like stochastic optimization method
\While{no stopping criterion is met}
    \State Shuffle $(\mathcal{X}_{P}, \mathcal{X}_{U})$ into $N$ mini-batches
    \For{$i=1$ to $N$}
        \If{$\tilde{\mathcal{R}}_{U}(g;\mathcal{X}_{U}^{i})-\pi_{P}\tilde{\mathcal{R}}_{P}(g;\mathcal{X}_{P}^{i}) \geq 0$}
            \State Update $\theta$ by $\mathcal{A}$ with $\nabla_{\theta}\tilde{\mathcal{R}}_{PU}(\mathcal{X}_{P}^{i},\mathcal{X}_{U}^{i})$
        \Else
            \State Update $\theta$ by $\mathcal{A}$ with $\nabla_{\theta}(\tilde{\mathcal{R}}_{U}(g;\mathcal{X}_{U}^{i}) - \pi \tilde{\mathcal{R}}_{P}(g;\mathcal{X}_{P}^{i}))$
        \EndIf
    \EndFor    
\EndWhile
\Returns{$\tilde{g}_{PU}(x;\theta)$}
\end{algorithmic}
\end{algorithm}

On a theoretical note, an important motivation for employing unbiased risk estimators in weakly supervised settings is that they enable estimation error bounds to guarantee statistical consistency. In \cite{Plessis2015}, for example, the authors prove that their risk estimator is both unbiased and consistent. On the other hand, proving consistency in the asymptotic cases is not informative for the behavior of the estimator in the finite-sample cases, especially when using complex models and deep learning methods \cite{Kiryo2017, Chou2020}. In combination with deep networks, unbiased risk estimators suffer from negative empirical risks during training, which is a sign of overfitting. Thus, biased risk estimators are often employed to control overfitting. The iFPU risk estimator proposed in this work relies on focal loss. Hence, the expected loss is not equal to the actual loss of the model, as focal loss is designed to assign different weights to different types of errors. This weighting can lead to an overestimate of the model's actual risk. Moreover, in the case of focal loss, it is not possible to guarantee that the expected loss will converge to the true risk. 

However, it is important to note that such a biased risk estimator can prove particularly useful in practical applications, where there is a need to improve the performance of the models on imbalanced datasets. In our case, the focal loss has been shown to be classification-calibrated \cite{FocalLossTheory2020}. Although it may give under-confident or over-confident classifiers, it can still yield the Bayes optimal classifier, which maximizes the expected accuracy in classification, i.e., it makes the most probable prediction for each new example. Finally, in terms of complexity, although the focal loss is more expensive than the logistic loss, since it requires logs and powers, in practice the overall asymptotic complexity remains the same.

\subsection{Theoretical analysis}\label{sec:thoery}
This section provides finite-sample guarantees and makes explicit which labeling assumptions are required. Proofs can be found in Appendix~\ref{app:proofs}. 

Let $\mathcal{G}$ be the score class, $p=\sigma(g(x))$, and write the focal loss of Eq.\ref{eq:FL} as $\ell_{+}(z)=-(1-\sigma(z))^{\gamma}\log\sigma(z)$ and $\ell_{-}(z)=\ell_{+}(-z)$. With partial risks $R_{P}^{+}=\mathbb{E}_{p_{P}}[\ell_{+}]$, $R_{P}^{-}=\mathbb{E}_{p_{P}}[\ell_{-}]$, $R_{U}^{-}=\mathbb{E}_{U}[\ell_{-}]$, the iFPU estimator is $\widehat{R}_{\mathrm{iFPU}}=\pi_{P}\widehat{R}_{P}^{+}-\pi_{P}\widehat{R}_{P}^{-}+\widehat{R}_{U}^{-}$, and the target is the focal risk $R_{\mathrm{FL}}(g)=\pi_{P}\mathbb{E}_{p_{P}}[\ell_{+}]+\pi_{N}\mathbb{E}_{N}[\ell_{-}]$, under full positive-negative supervision. 

To model labeling, let $e(x)=\Pr(s{=}1\mid y{=}1,x)$ be the propensity, $c=\mathbb{E}_{p_{P}}[e(x)]$, and $p_{l}=(e/c)\,p_{P}$ the labeled-positive density from which the labeled positives are drawn. In SCAR, we have $e(x)\equiv c$ (hence $p_{l}=p_{P}$), while SAR allows $e(\cdot)$ to depend on $x$.

The mixture identity $p=\pi_{P}p_{P}+\pi_{N}p_{N}$ used in Eqs.~(3)--(5) is a \emph{population} identity that holds for any loss and any labeling mechanism. The labeling assumption enters at the empirical positive average in Eqs.~(10)--(11), which estimates $\mathbb{E}_{p_{l}}[\cdot]$ rather than $\mathbb{E}_{p_{P}}[\cdot]$. Proposition~\ref{prop:bias} makes this precise.\\

\begin{proposition}[Population identity and SCAR validity versus SAR bias]
\label{prop:bias}
For any loss and any labeling, $R_{\mathrm{iFPU}}(g)=R_{\mathrm{FL}}(g)$
(population). The empirical estimator satisfies
\[
  \mathbb{E}\bigl[\widehat{R}_{\mathrm{iFPU}}(g)\bigr]=R_{\mathrm{FL}}(g)+b(g),
  \qquad
  b(g)=\pi_{P}\,\mathrm{Cov}_{p_{P}}\!\Bigl(\tfrac{e(x)}{c},\,
        \ell_{+}(g(x))-\ell_{-}(g(x))\Bigr).
\]
Hence $\widehat{R}_{\mathrm{iFPU}}$ is unbiased for $R_{\mathrm{FL}}$ under SCAR ($b\equiv0$), whereas under SAR, the bias $b(g)$ lives in the sampling mechanism of the labeled positives and is therefore orthogonal to the choice of loss.
\end{proposition}

For the SAR case, estimating $e(x)$ under PU data needs extra identifiability assumptions, as in the SAREM~\cite{Bekker2019} and LBE~\cite{LBE2022} approaches. Therefore, the SAR results of Section \ref{sec:results_and_discussion} are reported as \emph{empirical robustness} of the uncorrected estimator. This robustness is plausibly aided by the focal term up-weighting precisely the hard positives that SAR labeling down-weights in $b(g)$.

\paragraph{Estimation error under SCAR.}
The guarantees below target $R_{\mathrm{FL}}$ and therefore hold under SCAR.
 
\begin{lemma}[Focal loss is Lipschitz and bounded]
\label{lem:lip}
For all $\gamma\ge0$, $\ell_{\pm}$ are $L_{\gamma}$-Lipschitz in the score with $L_{\gamma}=1+\gamma/e$, and $0\le\ell_{\pm}\le C_{B}=-\sigma(B)^{\gamma}\log\sigma(-B)$ whenever $|g(x)|\le B$.
\end{lemma}

The lemma records two elementary properties of the focal loss. First, it is never too steep, since its slope is capped by $L_{\gamma}=1+\gamma/e$ and, second, on a bounded score range, it is never too tall, since it stays below $C_{B}$. The boundedness requires $|g(x)|\le B$ because the focal loss diverges for confidently wrong predictions ($p\to0$).

\begin{theorem}[SCAR estimation-error and excess-risk bound]
\label{thm:gen}
Under SCAR and $|g(x)|\le B$, with probability at least $1-\delta$ the following bound holds:
\[
  \sup_{g\in\mathcal{G}}\bigl|\widehat{R}_{\mathrm{iFPU}}(g)-R_{\mathrm{FL}}(g)\bigr|
  \le 2L_{\gamma}\bigl(2\pi_{P}\mathfrak{R}_{n_{P}}(\mathcal{G})
       +\mathfrak{R}_{n_{U}}(\mathcal{G})\bigr)
     +3C_{B}\sqrt{\tfrac{\log(6/\delta)}{2n_{\min}}}
  =\mathcal{O}\!\Bigl(\tfrac{1}{\sqrt{n_{P}}}+\tfrac{1}{\sqrt{n_{U}}}\Bigr),
\]
where $\mathfrak{R}_{n}$ is the Rademacher complexity and $n_{\min}=\min\{n_{P},n_{U}\}$. Consequently the iFPU minimiser $\widehat{g}$ has excess focal risk $R_{\mathrm{FL}}(\widehat{g})-\min_{g\in\mathcal{G}}R_{\mathrm{FL}}(g)$ of the same order.
\end{theorem}

The theorem certifies that the empirical PU risk iFPU actually minimizes is a faithful proxy for the population focal risk it cannot observe. Under SCAR the estimator is centered on that risk (Proposition~\ref{prop:bias}), and the gap between the two vanishes at the standard rate $\mathcal{O}(1/\sqrt{n_{P}}+1/\sqrt{n_{U}})$, so more labeled positives or more unlabeled data both tighten the guarantee. Since the bound holds for every $g$ at once, it transfers to the classifier iFPU returns, whose excess focal risk is of the same order.  Because focal loss is classification-calibrated~\cite{FocalLossTheory2020}, this excess \emph{surrogate} risk transfers to excess $0$--$1$ risk up to an approximation term, so the bound reaches the classification risk only through calibration. Recall that the biased focal estimator need not converge to the true risk as mentioned in Section \ref{sec:iFPU}.

\section{Experimental setup}\label{sec:experimental_setup}
In this section, we evaluate the effect of the proposed risk minimizer (iFPU) on the learning algorithm, comparing it with related PU learning methods under a common experimental setup. Rather than optimizing classifiers or hyperparameters, our focus is on assessing the impact of replacing other risk minimizers with iFPU. We also include comparisons with state-of-the-art PU methods that employ different approaches.

Since we focus on learning with PU data under class imbalance, without making explicit assumptions regarding the labeling mechanism, we experiment with both the SCAR and the SAR assumptions. In the experiments, we model the single training set scenario \cite{Bekker2020} and we use public datasets for imbalanced binary classification. In particular, we replicate the experimental setup of \cite{Ortega2023} to facilitate direct comparison against those methods. Since the datasets in their original form are not PU data, they are processed to (a) treat the negative examples as unlabeled data and (b) ``hide" part of the positive examples in the unlabeled data. 

The main contributions of the experimental setup are that it includes comparisons with multiple established PU methods and results under both SCAR and SAR assumptions, with SAR modeled to reflect the more challenging probabilistic gap scenario.

\subsection{Datasets}\label{sec:datasets}
As mentioned above, we use the same 14 datasets used in \cite{Ortega2023}, that cover different domains and are widely used in the imbalanced classification literature. We focus especially on such datasets since they provide a varying imbalance, spanning positive rates from about 1\% to 14\%. Table \ref{tab:datasets} summarizes the details of the datasets.

\begin{table}[ht]
\caption{Details of the 14 imbalanced classification datasets in terms of number of samples, features, imbalance ratio, and percentages of positive examples that become unlabeled to simulate PU data.}\label{tab:datasets}%
\begin{tabular}{@{}llllllll@{}}
\toprule
Dataset & Obs. & Feats & Pos. ratio & P75\% & P50\% & P25\% \\
\midrule
Cardio & 1831 & 21 & 9.60\% & 7.20\% & 4.80\% & 2.40\% \\
Thyroid & 3772 & 7 & 2.50\% & 1.87\% & 1.25\% & 0.63\% \\
Climate & 540 & 20 & 8.52\% & 6.39\% & 4.26\% & 2.13\% \\
Forest cover & 5000 & 10 & 1.00\% & 0.75\% & 0.50\% & 0.25\% \\
Seismic & 2584 & 11 & 6.57\% & 4.93\% & 3.29\% & 1.64\% \\
Mammography & 5000 & 6 & 2.58\% & 1.93\% & 1.29\% & 0.65\% \\
Shuttle & 5000 & 9 & 6.40\% & 4.80\% & 3.20\% & 1.60\% \\
Letter & 1600 & 32 & 6.25\% & 4.69\% & 3.13\% & 1.56\% \\
Yeast & 1004 & 8 & 9.86\% & 7.40\% & 4.93\% & 2.47\% \\
Poker 8-9vs5 & 2075 & 10 & 1.20\% & 0.90\% & 0.60\% & 0.30\% \\
Pie chart & 745 & 36 & 2.15\% & 0.61\% & 1.08\% & 0.54\% \\
Pizza cutter & 661 & 37 & 7.87\% & 5.90\% & 3.94\% & 1.97\% \\
Satellite & 5000 & 36 & 1.48\% & 1.11\% & 0.74\% & 0.37\% \\
Segment & 2308 & 18 & 14.0\% & 10.5\% & 7.0\% & 3.50\%\\
\bottomrule
\end{tabular}
\end{table}

In Table \ref{tab:datasets}, the column \emph{P} refers to the percentage of positive examples in the dataset, reflecting the different imbalance ratios in different datasets. As mentioned above, we treat negative examples as unlabeled and ``hide" positive examples in the unlabeled set. Thus, the columns \emph{P 25\%} to \emph{P 75\%} indicate the proportion of the positive examples that remain labeled in different experiments. This further demonstrates the effect of class imbalance, since the inherent imbalance between the positive and negative examples is gradually amplified by having fewer and fewer known positive ones during training at different settings.

In addition, the labeling mechanism decides which of the positive examples will become unlabeled. To simulate the SCAR assumption, each positive example has the same probability of becoming unlabeled. On the other hand, the SAR assumption introduces a bias to the selected positive examples that depends on the features. We use the average Euclidean distance between positive and negative examples to bias the labeling mechanism, as proposed in \cite{Ortega2023}. In particular, positive examples that are closer to negative ones in the original feature space are more likely to become unlabeled. 

The experiments are organized as follows. For each dataset, we perform 20 repetitions by splitting the data randomly into training (70\%) and test (30\%) sets. This is performed for both assumptions regarding the labeled positive examples (SCAR and SAR) and for all three ratios of positive examples that become unlabeled (25\%, 50\%, 75\%). The above combinations create a total of $(14 \, datasets \times 10 \, repetitions \times 2 \, labeling \, assumptions \times 3 \, ratios) = 840$ different runs of the methods.

\subsection{Methods and parameters}
We compare the proposed method\footnote{Code and data will be published on GitHub.} (iFPU) against three other methods that rely on empirical risk minimization, namely the unbiased PU (uPU) \cite{Plessis2015}, the non-negative PU (nnPU) \cite{Kiryo2017} and the imbalanced non-negative PU (i-NNPU) \cite{Su2021}, as neural instantiations commonly used in the literature, to assess the effect of the different empirical risks. We also include an XGBoost classifier for fairer comparison with the other approaches. Additionally, we compare iFPU against the two-step method (NNIF) proposed in \cite{Ortega2023}, which uses an anomaly detection step to create pseudo labels for the data and then learns an XGBoost classifier with default hyperparameters. For the latter comparison, we also implement an XGBoost classifier, equipped with the iFPU empirical risk estimator. 

Moreover, we compare our method against PU Hellinger Decision Trees (PUHDT) and PU Hellinger Random Forests (PUHRF) \cite{PUHellinger}, both known to achieve state-of-the-art performance in positive-unlabeled learning. Lastly, we also compare against the instance-dependent PU deep learning method with labeling bias estimation (LBE) \cite{LBE2022}, which specializes in hard settings with label bias, as in SAR, as well as with SAREM \cite{Bekker2019}, which is an Expectation-Maximization (EM) framework for PU learning and was one of the first approaches to explicitly tackle SAR. Additionally, we include results obtained from an XGBoost classifier treating the unlabeled instances as negatives (negativity assumption) as well as results from fully supervised training of the XGBoost classifier, serving as an upper bound for achievable classification performance. In this supervised scenario, we evaluate XGBoost using both the standard loss and the focal loss to further validate the suitability of focal loss under our class-imbalanced conditions. Table \ref{tab:models} summarizes all methods evaluated in our experiments.

\begin{table}[ht]
\caption{Methods used in empirical evaluation.}\label{tab:models}%
\begin{tabular}{@{}lll@{}}
\toprule
Method & Setting & Configuration\\
\midrule
uPU & Positive-Unlabeled & MLP, 5 layers, sigmoid loss \\
nnPU & Positive-Unlabeled & MLP, 5 layers, sigmoid loss \\
i-NNPU & Positive-Unlabeled & MLP, 5 layers, sigmoid loss \\
\textbf{iFPU$_{MLP}$} & Positive-Unlabeled & MLP, 5 layers, PU focal loss\\
\midrule
uPU$_{XGB}$ & Positive-Unlabeled & XGBoost, default hyper-parameters \\
nnPU$_{XGB}$ & Positive-Unlabeled & XGBoost, default hyper-parameters \\
NNIF & Positive-Unlabeled & XGBoost, default hyper-parameters \\
PUHDT & Positive-Unlabeled & PU Hellinger Decision Tree, default hyper-parameters \\
PUHRF & Positive-Unlabeled & PU Hellinger Random forest, default hyper-parameters \\
LBE & Positive-Unlabeled & MLP, default hyper-parameters \\
SAREM & Positive-Unlabeled & Expectation-Maximization, default parameters \\
\textbf{iFPU$_{XGB}$} & Positive-Unlabeled & XGBoost, PU focal loss, default hyper-parameters\\
\midrule
XGBoost & Positive-Unlabeled & Negativity assumption, standard loss, default hyper-parameters\\
XGBoost & Positive-Negative & XGBoost supervised, standard loss, default hyper-parameters\\
XGBoost$_{FL}$ & Positive-Negative & XGBoost supervised, focal loss, default hyper-parameters\\
\bottomrule
\end{tabular}
\end{table}

The focal loss introduces an additional parameter, $\gamma$. Specifically, the gradient of the focal loss is a weighted variant of the cross-entropy gradient, where the weighting depends on $\gamma$ through the term $p_t$. The latter is based on the model’s predictions and also appears in standard cross-entropy. The parameter $\gamma$ effectively controls the threshold probability at which the gradient flowing backward becomes small (or zero), indicating regions where the model is confident in its predictions. Hence, adjusting $\gamma$ allows the selection of a probability threshold below which predictions are considered non-confident and thus receive larger gradient updates. However, in this work we avoid hyper-parameter tuning, as it necessitates additional assumptions, especially under the SAR scenario, due to the absence of negatively labeled validation data \cite{Yoo2021}. Therefore, we adopt the recommended default setting of $\gamma = 3$.

\section{Empirical results}\label{sec:results_and_discussion}
We evaluate the performance of the methods on 14 heavily imbalanced datasets, with varying ratios of labeled positives under both SCAR and SAR labeling assumptions. Specifically, we investigate how our proposed risk estimator compares to related approaches based on similar risk-minimization frameworks, as well as to state-of-the-art methods. Although we experiment with two classifiers (MLP and XGBoost), our analysis primarily focuses on the impact of the risk minimizer rather than on the choice of classifier.

\begin{table}[ht]
\caption{Evaluation results under the SCAR assumption across different proportions of labeled positive examples. As shown in Table \ref{tab:models}, the first four models use an MLP classifier, while the next eight use state-of-the-art tree-based methods. In the final three rows, the first model operates under the negativity assumption, treating unlabeled examples as negative, and the other two are fully supervised and included for comparison purposes. Results are expressed in terms of ROC-AUC and PR-AUC for the positive class, macro-averaged over the 14 datasets. The best performing method is shown in bold and the runner up is underlined.}\label{tab:results_scar}
\tabcolsep=0.10cm
\begin{tabular*}{\textwidth}{@{\extracolsep\fill}lllllll}
\toprule%
& \multicolumn{3}{@{}c@{}}{ROC-AUC} & \multicolumn{3}{@{}c@{}}{PR-AUC} \\\cmidrule{2-4}\cmidrule{5-7}%
Labeled proportion: & 25\% & 50\% & 75\% & 25\% & 50\% & 75\% \\
\midrule
uPU  & 0.53$\pm$0.06 & 0.55$\pm$0.05 & 0.56$\pm$0.04 & 0.12$\pm$0.04 & 0.13$\pm$0.03 & 0.13$\pm$0.02 \\
nnPU & 0.54$\pm$0.06 & 0.55$\pm$0.05  & 0.56$\pm$0.04  & 0.12$\pm$0.04 & 0.13$\pm$0.02 & 0.13$\pm$0.02 \\
i-NNPU  & 0.54$\pm$0.05 & 0.56$\pm$0.05  & 0.56$\pm$0.03  & 0.12$\pm$0.03 & 0.13$\pm$0.02 & 0.13$\pm$0.02 \\
\textbf{iFPU$_{MLP}$} & 0.58$\pm$0.03 & 0.60$\pm$0.03  & 0.62$\pm$0.03 & 0.13$\pm$0.03 & 0.14$\pm$0.02 & 0.14$\pm$0.01 \\
\midrule
uPU$_{XGB}$ & 0.77$\pm$0.05 & 0.82$\pm$0.05 & 0.85$\pm$0.03 & 0.37$\pm$0.06 & 0.44$\pm$0.06 & 0.50$\pm$0.06 \\
nnPU$_{XGB}$ & 0.81$\pm$0.05 & 0.84$\pm$0.04 & 0.86$\pm$0.03 & 0.43$\pm$0.06 & 0.49$\pm$0.06 & 0.52$\pm$0.05 \\
NNIF  & \underline{0.81$\pm$0.05} & \underline{0.85$\pm$0.03} & \underline{0.87$\pm$0.03} & 0.46$\pm$0.06 & 0.50$\pm$0.06 & 0.53$\pm$0.05 \\
PUHDT & 0.72$\pm$0.09 & 0.77$\pm$0.07 & 0.79$\pm$0.07 & 0.40$\pm$0.07 & 0.43$\pm$0.06 & 0.46$\pm$0.05 \\
PUHRF & \textbf{0.83$\pm$0.07} & \textbf{0.87$\pm$0.06} & \textbf{0.89$\pm$0.06} & \textbf{0.52$\pm$0.09} & \textbf{0.59$\pm$0.07} & \textbf{0.63$\pm$0.06} \\
LBE & 0.80$\pm$0.06 & 0.84$\pm$0.05 & 0.86$\pm$0.04 & 0.49$\pm$0.08 & 0.53$\pm$0.06 & 0.53$\pm$0.06\\
SAREM & 0.81$\pm$0.05 & 0.84$\pm$0.04 & 0.85$\pm$0.04 & \underline{0.51$\pm$0.06} & 0.55$\pm$0.06 & 0.56$\pm$0.06 \\
\textbf{iFPU$_{XGB}$} & 0.80$\pm$0.04 & \underline{0.85$\pm$0.03}  & \underline{0.87$\pm$0.03} & \textbf{0.52$\pm$0.05} & \underline{0.57$\pm$0.04} & \underline{0.60$\pm$0.05} \\
\midrule
XGBoost$_{neg.}$ & 0.59$\pm$0.03 & 0.60$\pm$0.03  & 0.63$\pm$0.03 & 0.22$\pm$0.05 & 0.31$\pm$0.05 & 0.42$\pm$0.05 \\
XGBoost & 0.90$\pm$0.03 & 0.90$\pm$0.03  & 0.90$\pm$0.03 & 0.65$\pm$0.05 & 0.65$\pm$0.05 & 0.65$\pm$0.05 \\
XGBoost$_{FL}$ & 0.92$\pm$0.03 & 0.92$\pm$0.03  & 0.92$\pm$0.03 & 0.67$\pm$0.05 & 0.67$\pm$0.05 & 0.67$\pm$0.05 \\
\bottomrule
\end{tabular*}
\end{table}

Table \ref{tab:results_scar} presents the results under the SCAR assumption. In this setting, we include the methods of uPU \cite{Plessis2015}, nnPU \cite{Kiryo2017}, i-NNPU \cite{Su2021}, and our iFPU, using a 5-layer MLP classifier, as shown in the first four rows of the table. These methods demonstrate relatively low performance in terms of ROC-AUC and PR-AUC. The results of uPU, nnPU, and i-NNPU are comparable, with nnPU and i-NNPU offering little improvement over uPU, despite attempts to address overfitting \cite{Kiryo2017} or class imbalance \cite{Su2021}. iFPU outperforms the other three methods, albeit with a small margin.

The next eight rows of Table \ref{tab:results_scar} present a comparison between our proposed method, the two extensions of uPU and nnPU to an XGBoost classifier, and five state-of-the-art approaches: the two-step NNIF technique \cite{Ortega2023}, the PU Hellinger decision tree (PUHDT), and its ensemble variant (PUHRF) \cite{PUHellinger}, the instance-dependent PU with label bias (LBE) \cite{LBE2022}, and the SAREM approach \cite{Bekker2019}. For consistency, we employ the XGBoost classifier in our method, since NNIF utilizes XGBoost, and the Hellinger-based methods also rely on tree-based classifiers. The final three rows show the performance of the XGBoost classifier under a biased setting treating all unlabeled as negatives and under two fully supervised settings. The latter serves as an upper bound for PU classifiers. We report results using both the standard loss and the focal loss, clearly demonstrating the superior performance of the focal loss in this scenario and supporting our decision to minimize a risk estimator based on focal loss. It should be noted that the NNIF method includes an anomaly detection step designed to remove ``noisy" negative examples from the unlabeled set prior to training XGBoost with the remaining examples treated as negatives. In contrast, our iFPU method directly incorporates the proposed risk estimator into the XGBoost training process without requiring any pre-processing. Similarly, PUHDT, PUHRF, LBE, and SAREM integrate unlabeled examples directly during training without pre-processing steps. All tree-based methods use the same XGBoost and tree-based parameters to ensure a fair comparison.

The results confirm, as expected, that the number of labeled positive examples significantly influences both ROC-AUC and PR-AUC across all evaluated methods. Although ROC-AUC is reported, it should be noted that it is less reliable in scenarios of extreme class imbalance. This is because the false positive rate becomes insensitive to changes in the number of false positives, due to its denominator being dominated by a large number of true negatives. In contrast, PR-AUC is more sensitive to false positives and is unaffected by the number of true negatives, thus providing a more reliable metric in highly imbalanced contexts. Considering PR-AUC as the most suitable metric, our proposed iFPU method outperforms the uPU and nnPU extensions, as well as NNIF, PUHDT, LBE, and SAREM though it remains slightly below PUHRF. Additionally, aiming to measure significant performance differences among all methods of Table \ref{tab:results_scar}, the Iman-Davenport statistical test indeed reveals significant differences, ranking PUHRF first and iFPU second. However, to examine whether there are statistical differences between PUHRF and iFPU that achieve comparable performance, a subsequent Wilcoxon paired test indicates no statistically significant difference between these two across all three proportions of labeled positive examples (p-value $>$ 0.05).

\begin{table}[ht]
\caption{Evaluation results under the SAR assumption across different proportions of labeled positive examples. As shown in Table \ref{tab:models}, the first four models use an MLP classifier, while the next eight use state-of-the-art tree-based methods. In the final three rows, the first model operates under the negativity assumption, treating unlabeled examples as negative, and the other two are fully supervised and included for comparison purposes. Results are expressed in terms of ROC-AUC and PR-AUC for the positive class, macro-averaged over the 14 datasets. The best performing method is shown in bold and the runner up is underlined.}\label{tab:results_sar}
\tabcolsep=0.10cm
\begin{tabular*}{\textwidth}{@{\extracolsep\fill}lllllll}
\toprule%
& \multicolumn{3}{@{}c@{}}{ROC-AUC} & \multicolumn{3}{@{}c@{}}{PR-AUC} \\\cmidrule{2-4}\cmidrule{5-7}%
Labeled proportion: & 25\% & 50\% & 75\% & 25\% & 50\% & 75\% \\
\midrule
uPU  & 0.52$\pm$0.05 & 0.52$\pm$0.04 & 0.53$\pm$0.03 & 0.11$\pm$0.02 & 0.11$\pm$0.02 & 0.12$\pm$0.02 \\
nnPU & 0.53$\pm$0.05 & 0.53$\pm$0.04  & 0.54$\pm$0.03  & 0.11$\pm$0.02 & 0.11$\pm$0.02 & 0.12$\pm$0.02 \\
i-NNPU  & 0.54$\pm$0.05 & 0.53$\pm$0.03  & 0.54$\pm$0.02  & 0.12$\pm$0.02 & 0.12$\pm$0.02 & 0.12$\pm$0.02 \\
\textbf{iFPU$_{MLP}$} & 0.58$\pm$0.04 & 0.60$\pm$0.02  & 0.61$\pm$0.01 & 0.13$\pm$0.02 & 0.14$\pm$0.02 & 0.14$\pm$0.01 \\
\midrule
uPU$_{XGB}$ & 0.75$\pm$0.06 & 0.82$\pm$0.04 & 0.85$\pm$0.03 & 0.33$\pm$0.06 & 0.44$\pm$0.06 & 0.49$\pm$0.06 \\
nnPU$_{XGB}$ & 0.79$\pm$0.05 & 0.84$\pm$0.04 & 0.86$\pm$0.03 & 0.40$\pm$0.06 & 0.47$\pm$0.06 & 0.52$\pm$0.05 \\
NNIF  & \textbf{0.80$\pm$0.05} & \underline{0.85$\pm$0.03} & \underline{0.87$\pm$0.03} & 0.43$\pm$0.06 & 0.50$\pm$0.06 & 0.53$\pm$0.05 \\
PUHDT & 0.69$\pm$0.06 & 0.76$\pm$0.06 & 0.78$\pm$0.06 & 0.33$\pm$0.06 & 0.41$\pm$0.05 & 0.45$\pm$0.05 \\
PUHRF & \textbf0.80$\pm$0.06 & \textbf{0.86$\pm$0.06} & \textbf{0.89$\pm$0.05} & 0.45$\pm$0.06 & \textbf{0.57$\pm$0.06} & \textbf{0.62$\pm$0.06} \\
LBE & 0.79$\pm$0.05 & 0.84$\pm$0.04 & 0.85$\pm$0.04 & 0.47$\pm$0.07 & 0.53$\pm$0.06 & 0.53$\pm$0.06\\
SAREM & \textbf{0.80$\pm$0.05} & 0.84$\pm$0.04 & 0.85$\pm$0.04 & \underline{0.48$\pm$0.07} & 0.53$\pm$0.07 & 0.55$\pm$0.06\\
\textbf{iFPU$_{XGB}$} & \textbf{0.80$\pm$0.05} & 0.84$\pm$0.03  & \underline{0.87$\pm$0.05} & \textbf{0.50$\pm$0.05} & \underline{0.55$\pm$0.05} & \underline{0.60$\pm$0.05} \\
\midrule
XGBoost$_{neg.}$ & 0.58$\pm$0.03 & 0.60$\pm$0.03  & 0.63$\pm$0.03 & 0.18$\pm$0.05 & 0.28$\pm$0.05 & 0.40$\pm$0.05 \\
XGBoost & 0.90$\pm$0.03 & 0.90$\pm$0.03  & 0.90$\pm$0.03 & 0.65$\pm$0.05 & 0.65$\pm$0.05 & 0.65$\pm$0.05 \\
XGBoost$_{FL}$ & 0.92$\pm$0.03 & 0.92$\pm$0.03  & 0.92$\pm$0.03 & 0.67$\pm$0.05 & 0.67$\pm$0.05 & 0.67$\pm$0.05 \\
\bottomrule
\end{tabular*}
\end{table}

Under the more realistic SAR assumption (Table \ref{tab:results_sar}), the results differ notably. Again, the first four rows compare our proposed method with methods employing similar risk minimizers. Importantly, iFPU maintains performance comparable to that observed under the SCAR assumption (Table \ref{tab:results_scar}), whereas the performance of the other methods declines. This highlights the robustness of iFPU in scenarios where positive examples resembling negatives are less likely to be labeled. As anticipated, performance improves with an increasing percentage of labeled positive examples.

Examining the next eight rows of Table \ref{tab:results_sar} and again focusing on PR-AUC, we observe that all methods experience some performance degradation under the SAR assumption. However, NNIF, PUHRF, LBE, SAREM, and iFPU$_{XGB}$ exhibit greater resilience, with a smaller decrease in performance compared to PUHDT. Notably, iFPU outperforms all methods, except for PUHRF, across all three proportions of labeled positive examples. Specifically, iFPU achieves comparable results to PUHRF in the scenarios with 50\% and 75\% labeled positive examples. Importantly, in the most challenging scenario—where only 25\% of positive examples are labeled, iFPU surpasses PUHRF by 5 percentage points and LBE and SAREM by 3 and 2 percent points respectively. This supports the applicability of our proposed method in more realistic and challenging scenarios that frequently arise in real-world applications. In all settings, the biased XGBoost classifier under the negativity assumption is presented as an ablation of replacing the ``focused" risk estimator with the standard loss, highlighting the positive effect of using iFPU in an imbalanced PU setting. 

In terms of statistical significance across all methods, the Iman-Davenport statistical test reveals significant differences when considering all methods under the SAR assumption. Specifically, PUHRF ranks first, followed by iFPU, when 50\% or 75\% of the positive examples are labeled. In contrast, in the most challenging scenario with only 25\% labeled positives, iFPU ranks first. Furthermore, the Wilcoxon paired test indicates that iFPU is statistically indistinguishable from PUHRF at 50\% or 75\% labeled positive examples. However, in the most challenging setting with only 25\% labeled examples, iFPU significantly outperforms all the other methods (p-value $<$ 0.05).

\begin{figure}
\centering
\includegraphics[width=1\textwidth]{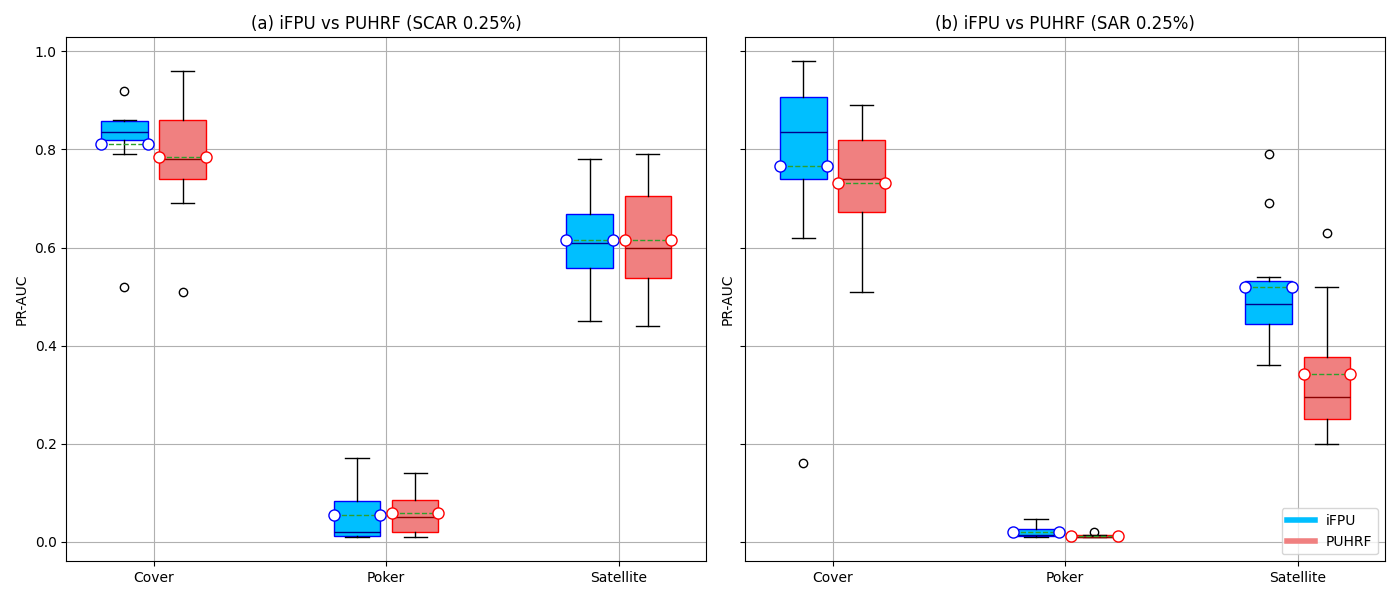}
\caption{Comparison of the iFPU$_{XGB}$ and PUHRF methods on the three most imbalanced datasets (Cover, Poker, Satellite) under (a) the SCAR assumption and (b) the SAR assumption, with only 25\% of positive examples labeled. Box plots summarize the PR-AUC performance across 10 independent runs for each method and dataset. The median is shown with a solid and the mean with a dotted line in each box.}\label{fig:boxplots}
\end{figure}

Analyzing the performance of the two best  methods (PUHRF and iFPU) further, Figure \ref{fig:boxplots} illustrates their performance on the three most heavily imbalanced datasets, considering the most challenging scenario - that is, when only 25\% of the positive examples are labeled - under both labeling assumptions. The box plots summarize results from ten independent runs of each method per dataset. In the relatively easier SCAR case, iFPU appears to achieve a slightly higher performance on the Cover dataset, exhibiting a higher mean and median PR-AUC values, whereas the two methods perform similarly on the other two datasets. Under the more challenging SAR assumption, iFPU clearly outperforms PUHRF in the Cover and Satellite datasets, as indicated by higher mean and median PR-AUC values, while the two methods achieve similar performance in the Poker dataset.

It is important to emphasize that no hyperparameter tuning was conducted in this study. While this was intentional, optimizing parameters for the XGBoost classifier and focal loss could potentially yield even stronger results. However, unbiased hyperparameter tuning in positive-unlabeled scenarios is challenging, as it requires additional assumptions due to the lack of negatively labeled validation data \cite{Yoo2021}. Therefore, addressing this challenge remains open for future research. It is also important to note that the focus of the controlled experiments was to show the effect of iFPU, keeping as much as possible similar classifiers with same hyper-parameters. Considering that the empirical evaluation was performed on tabular data, a natural consideration would be the use of TabPFN \cite{hollmann2025tabpfn}. Tabular foundation models are shown to perform remarkably well in a fully supervised positive-negative setting. Although iFPU is agnostic to the base classifier, a proper integration that minimises the iFPU risk over a pre-trained tabular representation is non-trivial because a frozen pre-trained model exposes no per-task empirical risk, and we flag it as future work at this point.

\subsection{Sensitivity analysis on class prior}
Accurate estimation of the class prior is essential for methods relying on prior probabilities, such as the approach proposed in this work. Although the prior $\pi_{p}$ can be effectively estimated from $(\mathcal{P})$, $(\mathcal{N})$, and $(\mathcal{U})$ data \cite{Saerens2002, Plessis2014b}, prior misspecification performance is crucial in such methods. In this section, we perform a sensitivity analysis to explore the impact of incorrect prior estimates on the performance of iFPU.

We use the same 14 datasets under the SCAR and SAR assumptions, focusing specifically on the most challenging scenario—when only 25\% of the positive examples are labeled. For this analysis, we evaluate the method using various prior estimates based on multiples of the ground-truth prior: $\pi_{P} \in \{ 0.25\pi_{P}, 0.5\pi_{P}, \pi_{P}, 1.5\pi_{P}, 2.0\pi_{P}, 4.0\pi_{P} \}$.

\begin{figure}
\centering
\includegraphics[width=0.9\textwidth]{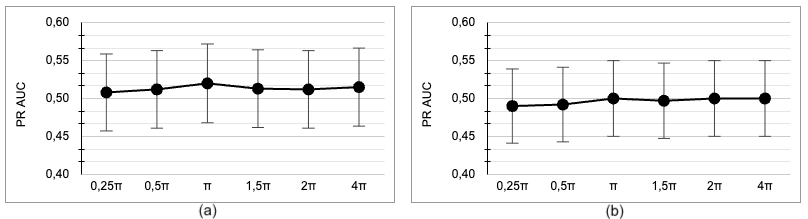}
\caption{Sensitivity analysis of the iFPU$_{XGB}$ method with respect to incorrect class prior estimates, averaged across all datasets. Performance is measured by macro-average PR-AUC of the positive class, under (a) the SCAR assumption and (b) the SAR assumption, with only 25\% of positive examples labeled. Error bars indicate standard deviations.}\label{fig:robust}
\end{figure}

Figure \ref{fig:robust} presents the results measured by PR-AUC under (a) the SCAR assumption and (b) the SAR assumption. We observe that iFPU generally exhibits robust performance across the entire range of incorrect prior estimates. The best results are achieved, as expected, when the correct prior is selected, whereas the largest performance degradations occur with significantly underestimated priors. Notably, the robustness of the method remains consistent regardless of the labeling assumption. 

\section{Real-world application}\label{sec:real_world}
Beyond the controlled experiments across benchmark datasets, we wanted to assess the value of the method in a real-world problem. For this purpose, we used it to detect financial misstatements in annual reports, a task where PU data naturally arise. Financial misstatements, which severely impact firms and stakeholders, demand accurate detection to ensure reports are error-free. This task is labor-intensive, involving the calculation of risk indicators and financial variables \cite{izavits2021, Bao2020, Bertomeu2021}. Thus, the literature relies on historical data to train ML models to predict the likelihood of future misstatements. The domain faces two major challenges: extreme class imbalance, as misstatements constitute a small fraction of the reports, and sparsely labeled positives, due to delayed identification of misstatements \cite{izavits2021, Bao2020}. Misstatements often remain undetected for years, affecting multiple reports. While historical data used for ML training are fully annotated, realistic modeling must account for the limited labeled positives and abundant unlabeled examples during training, reflecting real-world delays in detection.

Misstatements, whether intentional or due to accounting errors, are challenging to identify. Intentional misstatements are often deliberately concealed, while errors closely resemble normal cases, making detection difficult. This aligns with the probabilistic gap (PG) assumption, where unlabeled positives resemble negatives due to the labeling mechanism’s effort constraints. Thus, financial misstatement detection exemplifies learning from imbalanced PU data, under the PG assumption.

\subsection{Data and experimental setting}\label{sec:misstatement_setting}
We used the data and experimental setup from \cite{izavits2021, izavitsESWA}, based on financial information from \cite{Bao2020}, which includes publicly traded US companies and consolidated financial statements. Each record represents a financial report for a specific year, with 28 financial indices, such as assets, debt, receivables, liabilities, and 14 derived features from the work in \cite{Dechow2011}, such as soft assets, earning before interest, and market data which have proven effective in misstatement detection \cite{, Bao2020, Bertomeu2021}. Additional categorical features, such as industry and state of incorporation, are also included. For details on the financial variables, the reader is referred to \cite{Cecchini2010, Bao2020} and Appendix \ref{app:misstatement}. The data spans the period 2000–2014, aligned with the test years of \cite{Bao2020, izavits2021, izavitsESWA, izavits2025}.

The dataset is structured with test years from 2003–2014, and for each test year, the preceding three years are used for training (e.g., 2000–2002 for 2003). In total, there are 47,086 firm-year instances in the data. Each instance in the dataset is labeled as positive (misstatement) or negative, with positives constituting only 1\% of the dataset. While all labels are available, this does not reflect reality: misstatements are often discovered years after filing, and may retroactively affect multiple prior reports \cite{Dyck2010}. Empirical evidence suggests that overlooking this delay can significantly bias model performance \cite{izavits2021}.

\begin{figure}
  \centering
  \includegraphics[width=\linewidth]{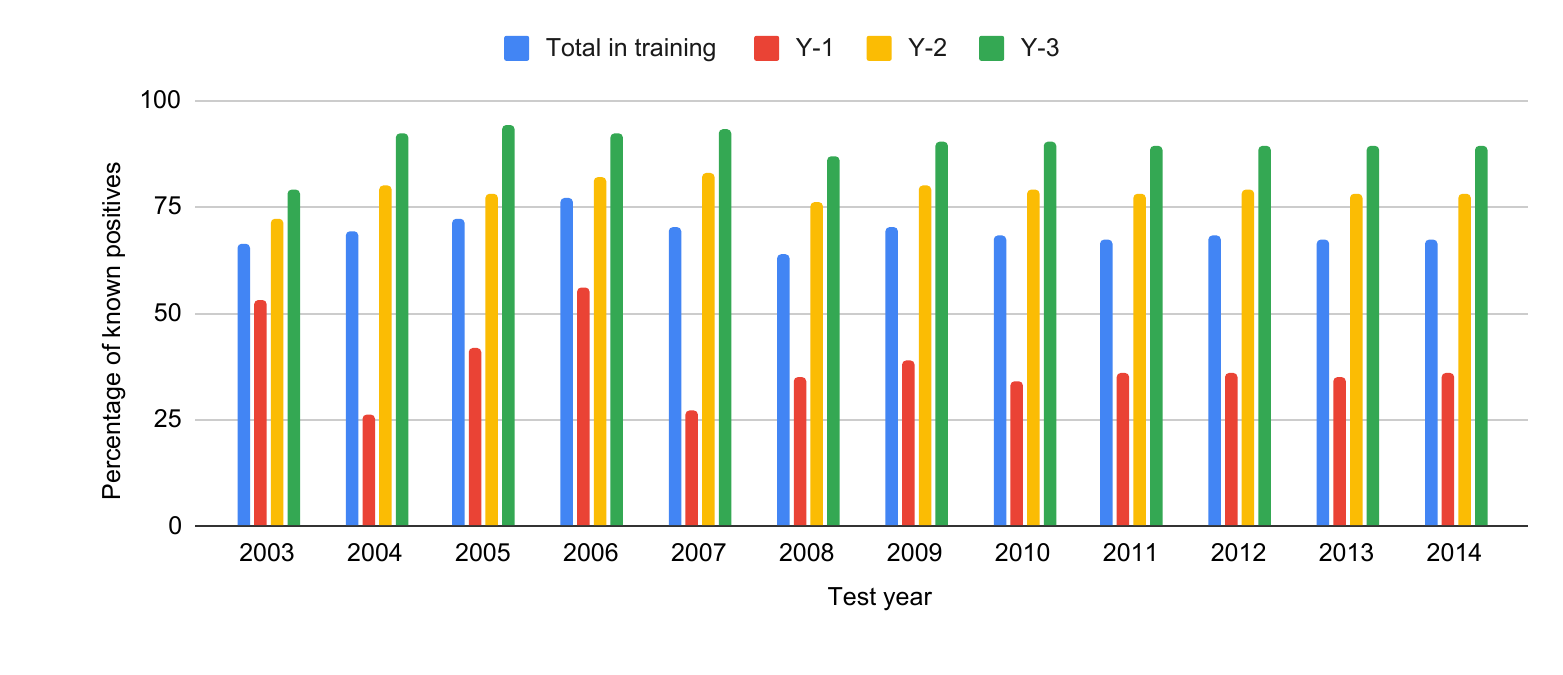}
  \caption{The percentage of the positive examples (misstatements) that are known per training year. \emph{Y-1}, \emph{Y-2}, and \emph{Y-3} correspond to the three years before the test year.}
  \label{fig:pos_percent_each_year}
\end{figure}

To simulate the delayed detection of financial misstatements, labels in the training set are flipped for ``unknown positives'' if their restatement date falls after the test year, based on the Audit Analytics (AA) database\footnote{\url{https://www.auditanalytics.com/}}. This ensures a realistic setup with hidden positives in the training data. Following this approach, as shown in Figure \ref{fig:pos_percent_each_year}, approximately 40\% of positive training labels are missing at the time of training, with higher percentages for years closer to the test year. For example, when testing on 2003, only 50\% of positives are retained for 2002 (\emph{Y-1} year), 70\% for 2001 (\emph{Y-2} year), and 80\% for 2000 (\emph{Y-3} year), resulting in about 40\% of positives hidden in the unlabeled pool (blue bar).

We adopt the model architecture from \cite{izavitsESWA}, modifying it to incorporate the proposed empirical risk. The model features a TabTransformer \cite{tabtransformer2020} for contextual embeddings of categorical attributes, which enhances predictive accuracy by capturing interactions between categorical variables. These embeddings are concatenated with continuous attributes to form a final feature vector, which is processed by a gated MLP (gMLP) module \cite{gmlp2021}. The gMLP encodes cross-feature interactions using a gating mechanism similar to attention. To address class imbalance and detect overlapping positives and negatives, a calibrated focal loss is minimized. We adapt this by incorporating the proposed iFPU method, which uses the same calibration mechanism for focal loss as \cite{izavitsESWA}. Specifically, the focal loss parameter $\gamma$ is set to 3 or 5, based on model confidence, and $\alpha$ is set to the inverse class frequency. To compare the PU setting against prior work, we use the same hyperparameters as \cite{izavitsESWA} without additional tuning. Moreover, we include the iFPU$_{XGB}$ method described in this work, i.e., the proposed iFPU risk estimator with the XGBoost classifier, as well as the PUHRF method, which was also included in our controlled experiments of Section \ref{sec:results_and_discussion}.

\subsection{Empirical results}\label{sec:misstatement_evaluation}
In the setting described in section \ref{sec:misstatement_setting}, we evaluate the classification models as rankers, similar to previous work \cite{izavits2021, Bao2020, izavitsESWA, izavits2025}, since the task requires associating every report with a score, i.e., the classifier’s confidence. In addition, it has been argued \cite{Bao2020, Bertomeu2021} that the time capacity of human auditors is limited, and, in practice, auditors can choose to see as many high-risk reports as possible. Thus, we report results on R-precision, which is a ranking evaluation measure for direct comparison with the previous work.

We compare the method that incorporates the calibrated-iFPU and the proposed iFPU$_{XGB}$, as well as the PUHRF \cite{PUHellinger}, against the work in \cite{izavits2021}, which uses the TabTransformer with the gMLP and the calibrated focal loss (TabTransformer\_gMLP\_Cal\_FL) and we also include the methods in \cite{izavits2021} (Fin-SVM and Fin-LR), the RUSBoost from \cite{Bao2020, Bertomeu2021}, and the methodology proposed in \cite{Bao2020}, where there is a temporal gap of two years between training and testing (SVM-2y gap and LR-2y gap). The assumption behind this two-year gap is that most misstatements will have been detected meanwhile, and thus, a PU setting may not be required. Our exploratory analysis confirmed that these two years contain the majority of unknown positives (see Fig. \ref{fig:pos_percent_each_year}). However, unknown positives exist also beyond the two-year period.

\begin{table}[h]
\caption{Performance of different methods in terms of R-precision, averaged over the period 2003-2014.}
  \label{tab:average_performance}
\begin{tabular}{@{}lr@{}}
\toprule
Method & R-precision\\
\midrule
LR-2y gap from \cite{Bao2020} & 2.40\% \\
SVM-2y gap from \cite{Bao2020} & 2.00\% \\
RUSBoost from \cite{Bao2020, Bertomeu2021} & 5.73\% \\
Fin-SVM from \cite{izavits2021} & 3.40\% \\
Fin-LR from \cite{izavits2021} & 4.70\% \\
TabTransformer\_gMLP\_Cal\_FL from \cite{izavitsESWA} & 17.63\% \\
\midrule
PUHRF & 18.22\% \\
Calibrated-iFPU & \underline{19.10\%} \\
iFPU$_{XGB}$ & \textbf{19.95\%} \\
\bottomrule
\end{tabular}
\end{table}

In terms of empirical results in the context of this task, it has been argued \cite{Bao2020, izavits2021, Bertomeu2021} that it is common for suggested models to perform poorly since the challenges of the task and the peculiarities of the financial domain do not allow much space for generalization. However, as shown in Table \ref{tab:average_performance}, the TabTransformer model with the calibrated focal loss managed to achieve an R-precision score three times higher than the RUSBoost model. An important note to make here is that all the methods presented in Table \ref{tab:average_performance}, except the ones using iFPU or PUHRF, do not assume a PU setting. This means that although they are all evaluated on the same data, using the same number of positive examples in the training sets, they rely on the negativity assumption, treating all unlabeled examples as negative. On the other hand, when we model the problem as a PU learning task, we achieve higher performance, as shown in the last two rows of the table, which, to the best of our knowledge, constitutes the state-of-the-art in misstatement detection. Notably, in this realistic PU setting, the methods incorporating iFPU perform better than PUHRF, indicating that the proposed method is more suitable for handling real-world cases where PU data resembling the probabilistic gap assumption arises.

\section{Conclusions}\label{sec:conclusion}
While PU learning has been widely studied, its application to imbalanced PU data remains underexplored. We address this gap by proposing a PU learning method based on the risk minimization framework, incorporating focal loss as a surrogate loss function. This approach uses both positive and unlabeled examples for training, leveraging the focal loss to emphasize hard-to-identify positive examples. Unlike two-step PU learning methods, the proposed approach integrates seamlessly with any classification algorithm based on cost minimization.

We extensively evaluated the method on 14 benchmark datasets and a real-world application, simulating various labeling mechanisms to reflect natural PU data emergence. The method’s performance highlights the potential of risk minimization techniques for PU learning. At the same time, extensions to semi-supervised and multi-class settings are natural directions for future research.

Other future directions include testing on additional datasets, including more real-world scenarios, and extending the method to multi-positive and unlabeled settings. Additionally, temperature scaling could be explored to calibrate model outputs toward true class posteriors, though optimizing the required temperature parameter in PU data presents a non-trivial challenge that is also worth exploring.

\backmatter


\begin{appendices}

\section{Proofs}\label{app:proofs}
Throughout, $\mathfrak{R}_{n}(\mathcal{G})=\mathbb{E}\bigl[\sup_{g\in\mathcal{G}}\tfrac{1}{n}\sum_{i=1}^{n}\varepsilon_{i}\,g(x_{i})\bigr]$ denotes the Rademacher complexity of $\mathcal{G}$, with $\varepsilon_{i}$ i.i.d.\ Rademacher signs. We use $\ell_{\pm}$, $R_{P}^{\pm}$, $R_{U}^{-}$, $R_{\mathrm{FL}}$, $e(\cdot)$, $c$, and $p_{l}=(e/c)p_{P}$ as defined in Section \ref{sec:thoery}.
 
\subsection{Proof of Proposition~\ref{prop:bias}}
 
\begin{proof}[Proof of Proposition~\ref{prop:bias}]
\emph{Population identity.}
The mixture identity $p(x)=\pi_{P}p_{P}(x)+\pi_{N}p_{N}(x)$ gives, for any fixed
$h$, $\mathbb{E}_{U}[h]=\pi_{P}\mathbb{E}_{p_{P}}[h]+\pi_{N}\mathbb{E}_{N}[h]$.
Taking $h=\ell_{-}(g(\cdot))$ yields
$\pi_{N}R_{N}^{-}(g)=R_{U}^{-}(g)-\pi_{P}R_{P}^{-}(g)$, i.e.\ Eq.~(3).
Substituting into
$R_{\mathrm{FL}}(g)=\pi_{P}R_{P}^{+}(g)+\pi_{N}R_{N}^{-}(g)$ gives
$R_{\mathrm{FL}}(g)=\pi_{P}R_{P}^{+}(g)-\pi_{P}R_{P}^{-}(g)+R_{U}^{-}(g)
=R_{\mathrm{iFPU}}(g)$, using only the mixture identity, hence for any loss and
any labeling mechanism.
 
\emph{Bias of the empirical estimator.}
The labeled positives are drawn from $p_{l}$, so the positive averages have mean
$\mathbb{E}_{p_{l}}[\cdot]$ while the unlabeled average has mean
$\mathbb{E}_{U}[\cdot]$. By linearity and the population identity,
\[
  \mathbb{E}\bigl[\widehat{R}_{\mathrm{iFPU}}(g)\bigr]-R_{\mathrm{FL}}(g)
  =\pi_{P}\bigl(\mathbb{E}_{p_{l}}-\mathbb{E}_{p_{P}}\bigr)\!\bigl[\ell_{+}-\ell_{-}\bigr],
\]
since the unlabeled term is unchanged. From $p_{l}=(e/c)p_{P}$,
\[
  \bigl(\mathbb{E}_{p_{l}}-\mathbb{E}_{p_{P}}\bigr)[h]
  =\mathbb{E}_{p_{P}}\!\Bigl[\bigl(\tfrac{e(x)}{c}-1\bigr)h\Bigr],
  \qquad h=\ell_{+}-\ell_{-}.
\]
Because $\mathbb{E}_{p_{P}}[e(x)/c]=1$, the multiplier $e(x)/c-1$ is mean-zero
under $p_{P}$, so this expectation equals
$\mathrm{Cov}_{p_{P}}\!\bigl(e(x)/c,\,h\bigr)$. This proves the stated form of
$b(g)$. Under SCAR, $e(x)\equiv c$, so $e(x)/c$ is constant and the covariance
vanishes for every $g$; under SAR it is generally nonzero. The multiplier
depends only on the labeling mechanism and not on $\ell$, establishing that the bias
is orthogonal to the choice of loss.
\end{proof}

\subsection{Proof of Lemma~\ref{lem:lip}}
 
\begin{proof}[Proof of Lemma~\ref{lem:lip}]
Since $\ell_{-}(z)=\ell_{+}(-z)$, it suffices to bound $|\ell_{+}'(z)|$. We write
$p=\sigma(z)$, and use sigmoid's derivative $\mathrm{d}p/\mathrm{d}z=p(1-p)$. Differentiating
$\ell_{+}=-(1-p)^{\gamma}\log p$,
\[
  \frac{\mathrm{d}\ell_{+}}{\mathrm{d}p}
  =\gamma(1-p)^{\gamma-1}\log p-\frac{(1-p)^{\gamma}}{p},
\]
hence, by the chain rule,
\begin{equation}
  \ell_{+}'(z)=\frac{\mathrm{d}\ell_{+}}{\mathrm{d}p}\,p(1-p)
  =\gamma\,p\,(1-p)^{\gamma}\log p-(1-p)^{\gamma+1}.
  \label{eq:focal-deriv-app}
\end{equation}
Both terms in \eqref{eq:focal-deriv-app} are non-positive on $(0,1)$, so
$|\ell_{+}'(z)|=\gamma\,p\,|\log p|\,(1-p)^{\gamma}+(1-p)^{\gamma+1}$.
The map $p\mapsto -p\log p$ on $(0,1)$ is maximised at $p=1/e$ with value $1/e$;
with $(1-p)^{\gamma}\le1$ this bounds the first term by $\gamma/e$, and the
second by $1$. Thus $|\ell_{+}'(z)|\le 1+\gamma/e=L_{\gamma}$.
 
For boundedness, \eqref{eq:focal-deriv-app} shows $\ell_{+}'(z)\le0$, so
$\ell_{+}$ is non-increasing and attains its maximum on $[-B,B]$ at $z=-B$,
where $p=\sigma(-B)$ and $1-p=\sigma(B)$, giving
$\ell_{+}(-B)=-\sigma(B)^{\gamma}\log\sigma(-B)=C_{B}$; non-negativity follows
from $-\log p\ge0$. Symmetry gives the same bounds for $\ell_{-}$.
\end{proof}

\subsection{Proof of Theorem~\ref{thm:gen}}
 
\begin{proof}[Proof of Theorem~\ref{thm:gen}]
Under SCAR, $p_{l}=p_{P}$, so by Proposition~\ref{prop:bias} the empirical
positive averages are unbiased for $R_{P}^{\pm}$ and
$\mathbb{E}[\widehat{R}_{\mathrm{iFPU}}]=R_{\mathrm{FL}}$. The triangle
inequality gives
\[
  \sup_{g}\bigl|\widehat{R}_{\mathrm{iFPU}}-R_{\mathrm{FL}}\bigr|
  \le\pi_{P}\sup_{g}\bigl|\widehat{R}_{P}^{+}-R_{P}^{+}\bigr|
    +\pi_{P}\sup_{g}\bigl|\widehat{R}_{P}^{-}-R_{P}^{-}\bigr|
    +\sup_{g}\bigl|\widehat{R}_{U}^{-}-R_{U}^{-}\bigr|.
\]
Fix one term, an average of $n$ i.i.d.\ terms of the form $\ell(g(x))$ bounded
by $C_{B}$ (Lemma~\ref{lem:lip}). Changing one sample shifts the supremum by at
most $C_{B}/n$, so McDiarmid's inequality gives, with probability at least
$1-\delta/3$,
\[
  \sup_{g}\bigl|\widehat{R}-R\bigr|
  \le \mathbb{E}\sup_{g}\bigl|\widehat{R}-R\bigr|
      +C_{B}\sqrt{\tfrac{\log(6/\delta)}{2n}}.
\]
Standard symmetrisation bounds the expectation by
$2\,\mathfrak{R}_{n}(\ell\circ\mathcal{G})$, and the Ledoux-Talagrand
contraction inequality \cite{foster2019ellinftyvectorcontractionrademacher, Bartlett2003},
applicable since $\ell$ is $L_{\gamma}$-Lipschitz (Lemma~\ref{lem:lip}), gives
$\mathfrak{R}_{n}(\ell\circ\mathcal{G})\le L_{\gamma}\mathfrak{R}_{n}(\mathcal{G})$.
Hence each term is at most
$2L_{\gamma}\mathfrak{R}_{n}(\mathcal{G})+C_{B}\sqrt{\log(6/\delta)/(2n)}$
with $n=n_{P}$ for the two positive terms and $n=n_{U}$ for the unlabeled term.
A union bound over the three events, together with $\pi_{P}\le1$,
$2\pi_{P}+1\le3$, and $n\ge n_{\min}$, yields the displayed bound. The order
statement follows when $\mathfrak{R}_{n}(\mathcal{G})=\mathcal{O}(1/\sqrt{n})$,
as for norm-bounded linear or neural classes.
 
\emph{Excess risk.}
Let $\Delta=\sup_{g}|\widehat{R}_{\mathrm{iFPU}}(g)-R_{\mathrm{FL}}(g)|$,
$\widehat{g}\in\arg\min_{g}\widehat{R}_{\mathrm{iFPU}}(g)$ and
$g^{\star}\in\arg\min_{g}R_{\mathrm{FL}}(g)$. Then
$R_{\mathrm{FL}}(\widehat{g})-R_{\mathrm{FL}}(g^{\star})
\le(\widehat{R}_{\mathrm{iFPU}}(\widehat{g})+\Delta)
-(\widehat{R}_{\mathrm{iFPU}}(g^{\star})-\Delta)\le2\Delta$,
since $\widehat{R}_{\mathrm{iFPU}}(\widehat{g})\le\widehat{R}_{\mathrm{iFPU}}(g^{\star})$.
Thus the excess focal risk is of the same order as the bound above.
\end{proof}

\section{Financial features for misstatement detection}\label{app:misstatement}
This section provides the list of financial features used for the real-world modeling of misstatement detection.

\begin{table}[ht]
  \caption{Financial feature set from \cite{izavits2021}, based on ~\cite{Bao2020} and~\cite{Dechow2011}.}
  \label{tab:fin_features}
  \begin{tabular}{cc}
    \toprule
 Current assets, total (act) & Account payable (ap) \\
 Assets, total (at) & Common/ordinary equity, total (ceq) \\
Cash \& short investments (che) & Cost of goods sold (cogs)\\
 Common shares outstanding (csho) & Debt in current liabilities, total (dlc) \\
 Inventories, total (invt) & Long-term debt issuance (dltis)\\
 Long-term debt, total (dltt) & Net income (ni) \\
Depreciation \& amortization (dp) & Receivables, total (rect)\\
 Income bf extraordinary items (ib) & Investment and advances, other (ivao) \\
 Retained earnings (re) & Short-term inv/ments, total (ivst) \\
 Current liabilities, total (lct) & Liabilities, total (lt)\\
 Preferred stock (capital), total (pstk) & Sales/turnover (sale) \\
 Sale of common/preferred stock (sstk) & Income taxes payable (txp) \\
Income taxes, total (txt) & Interest and expense, total (xint)\\
 Price close, annual, fiscal (prcc\_f) & WC accruals (dch\_wc) \\
 RSST accruals (ch\_rsst) & Change in receivables (dch\_rec) \\
 Change in inventory (dch\_inv) & Change in cash sales (ch\_cs)\\
 Soft assets percentage (soft\_assets) & Change in cash margin (ch\_cm) \\
 Change in return on assets (ch\_roa) & actual issuance (issue) \\
 Book to market (bm) & Depreciation index (dpi)\\
 Retained earnings over assets (reoa) & Earnings before interest (EBIT) \\
 Cash Flow Earnings Difference (ch\_fcf) & Property \& equipment, total (ppegt) \\
\bottomrule
  \end{tabular}
\end{table}

\end{appendices}

\bibliography{bibliography}

@String{Computer = "{IEEE} Computer" }

@String{Springer = "Springer-Verlag" }

@article{Zhou2017,
    author = {Zhou, Zhi-Hua},
    title = {A brief introduction to weakly supervised learning},
    journal = {National Science Review},
    volume = {5},
    number = {1},
    pages = {44-53},
    year = {2017},
    month = {08},
    doi = {10.1093/nsr/nwx106}
}

@article{Frenay2014,
  author={Frenay, Benoit and Verleysen, Michel},
  journal={IEEE Transactions on Neural Networks and Learning Systems}, 
  title={Classification in the Presence of Label Noise: A Survey}, 
  year={2014},
  volume={25},
  number={5},
  pages={845-869},
  doi={10.1109/TNNLS.2013.2292894}
}

@article{Bekker2020,
	author = {Bekker, Jessa and Davis, Jesse},
	date = {2020/04/01},
	doi = {10.1007/s10994-020-05877-5},
	id = {Bekker2020},
	isbn = {1573-0565},
	journal = {Machine Learning},
	number = {4},
	pages = {719--760},
	title = {Learning from positive and unlabeled data: a survey},
	url = {https://doi.org/10.1007/s10994-020-05877-5},
	volume = {109},
	year = {2020}
}

@inproceedings{Bekker2019,
author = {Bekker, Jessa and Robberechts, Pieter and Davis, Jesse},
title = {Beyond the Selected Completely at Random Assumption for Learning from Positive and Unlabeled Data},
year = {2019},
isbn = {978-3-030-46146-1},
doi = {https://doi.org/10.1007/978-3-030-46147-8\_5},
booktitle = {Machine Learning and Knowledge Discovery in Databases: European Conference, ECML PKDD},
numpages = {71–85},
publisher = {Springer},
address = {Wurzburg, Germany}
}

@article{Claesen2015,
author = {Claesen, Marc and De Smet, Frank and Suykens, Johan and De Moor, Bart},
year = {2015},
month = {07},
pages = {73-84},
title = {A Robust Ensemble Approach to Learn From Positive and Unlabeled Data Using SVM Base Models},
volume = {160},
journal = {Neurocomputing},
doi = {10.1016/j.neucom.2014.10.081}
}

@article{Stripling2018,
title = {Isolation-based conditional anomaly detection on mixed-attribute data to uncover workers' compensation fraud},
journal = {Decision Support Systems},
volume = {111},
pages = {13-26},
year = {2018},
issn = {0167-9236},
doi = {https://doi.org/10.1016/j.dss.2018.04.001},
url = {https://www.sciencedirect.com/science/article/pii/S016792361830068X},
author = {Eugen Stripling and Bart Baesens and Barak Chizi and Seppe {vanden Broucke}},
}

@inproceedings{izavits2021,
author = {Zavitsanos, Elias and Mavroeidis, Dimitris and Bougiatiotis, Konstantinos and Spyropoulou, Eirini and Loukas, Lefteris and Paliouras, Georgios},
title = {Financial Misstatement Detection: A Realistic Evaluation},
year = {2021},
doi = {10.1145/3490354.3494453},
booktitle = {Proceedings of the Second ACM International Conference on AI in Finance},
articleno = {34},
numpages = {9},
address = {Virtual event},
publisher = {ACM}
}

@article{Mignone2020a,
	author = {Mignone, Paolo and Pio, Gianvito and D{\v z}eroski, Sa{\v s}o and Ceci, Michelangelo},
	doi = {10.1038/s41598-020-78033-7},
	journal = {Scientific Reports},
	number = {1},
	pages = {22295},
	title = {Multi-task learning for the simultaneous reconstruction of the human and mouse gene regulatory networks},
	url = {https://doi.org/10.1038/s41598-020-78033-7},
	volume = {10},
	year = {2020}
}

@article{Mignone2020b,
    author = {Mignone, Paolo and Pio, Gianvito and D’Elia, Domenica and Ceci, Michelangelo},
    title = "{Exploiting transfer learning for the reconstruction of the human gene regulatory network}",
    journal = {Bioinformatics},
    volume = {36},
    number = {5},
    pages = {1553-1561},
    year = {2019},
    month = {10},
    doi = {10.1093/bioinformatics/btz781}
}

@article{Vasighizaker2018,
title = {C-PUGP: A cluster-based positive unlabeled learning method for disease gene prediction and prioritization},
journal = {Computational Biology and Chemistry},
volume = {76},
pages = {23-31},
year = {2018},
issn = {1476-9271},
doi = {https://doi.org/10.1016/j.compbiolchem.2018.05.022},
url = {https://www.sciencedirect.com/science/article/pii/S1476927117306485},
author = {Akram Vasighizaker and Saeed Jalili},
}

@article{Claesen2015b,
author = {Claesen, Marc and De Smet, Frank and Gillard, Pieter and Mathieu, Chantal and De Moor, Bart},
year = {2015},
month = {04},
pages = {},
title = {Building Classifiers to Predict the Start of Glucose-Lowering Pharmacotherapy Using Belgian Health Expenditure Data},
journal = {ArXiv},
volume={abs/1504.07389},
}

@article{Galarrage2015,
author = {Gal\'{a}rraga, Luis and Teflioudi, Christina and Hose, Katja and Suchanek, Fabian M.},
title = {Fast Rule Mining in Ontological Knowledge Bases with AMIE+},
year = {2015},
issue_date = {December  2015},
publisher = {Springer-Verlag},
address = {Berlin, Heidelberg},
volume = {24},
number = {6},
issn = {1066-8888},
url = {https://doi.org/10.1007/s00778-015-0394-1},
doi = {10.1007/s00778-015-0394-1},
journal = {The VLDB Journal},
month = {dec},
pages = {707–730},
numpages = {24}
}

@inproceedings{Zupanc2018,
author = {Zupanc, Kaja and Davis, Jesse},
title = {Estimating Rule Quality for Knowledge Base Completion with the Relationship between Coverage Assumption},
year = {2018},
doi = {10.1145/3178876.3186006},
booktitle = {Proceedings of the 2018 World Wide Web Conference},
pages = {1073–1081},
address = {Lyon, France},
publisher = {International World Wide Web Conferences Steering Committee}
}

@InProceedings{Japkowicz2000,
author = {Japkowicz, Nathalie},
year = {2000},
month = {06},
pages = {},
title = {The Class Imbalance Problem: Significance and Strategies},
booktitle = {Proceedings of the 2000 International Conference on Artificial Intelligence ICAI},
address = {Athens, Georgia, USA},
publisher = {Computer Science Research, Education and Applications Press}
}

@article{Krawczyk2016,
	author = {Krawczyk, Bartosz},
	date = {2016/11/01},
	doi = {10.1007/s13748-016-0094-0},
	id = {Krawczyk2016},
	isbn = {2192-6360},
	journal = {Progress in Artificial Intelligence},
	number = {4},
	pages = {221--232},
	title = {Learning from imbalanced data: open challenges and future directions},
	url = {https://doi.org/10.1007/s13748-016-0094-0},
	volume = {5},
	year = {2016}
}

@article{Fung2006,
  author={Gabriel Pui Cheong Fung and Yu, J.X. and Hongjun Lu and Yu, P.S.},
  journal={IEEE Transactions on Knowledge and Data Engineering}, 
  title={Text classification without negative examples revisit}, 
  year={2006},
  volume={18},
  number={1},
  pages={6-20},
  doi={10.1109/TKDE.2006.16}
}

@inproceedings{Liu2003,
author = {Liu, Bing and Yu, Philip and Li, Xiaoli},
year = {2003},
title = {Partially Supervised Classification of Text Documents},
booktitle = {International Conference on Machine Learning (ICML)},
doi = {10.1385/1-59259-358-5:387},
publisher = {AAAI press},
address = {Washington, DC USA},
pages = {8}
}

@inproceedings{Yu2002,
author = {Yu, Hwanjo and Han, Jiawei and Chang, Kevin Chen-Chuan},
title = {PEBL: Positive Example Based Learning for Web Page Classification Using SVM},
year = {2002},
doi = {10.1145/775047.775083},
booktitle = {8th ACM SIGKDD International Conference on Knowledge Discovery and Data Mining},
pages = {239–248},
numpages = {10},
address = {	Edmonton Alberta Canada},
publisher = {ACM}
}

@article{Yu2003,
author = {Yu, Hwanjo and Han, Jiawei and Chang, Kevin},
year = {2003},
month = {10},
pages = {},
title = {PEBL: Web Page Classification without Negative Examples},
volume = {16},
journal = {IEEE Transactions on Knowledge and Data Engineering},
doi = {10.1109/TKDE.2004.1264823}
}

@article{Peng2008,
	author = {Peng, Tao and Zuo, Wanli and He, Fengling},
	date = {2008/09/01},
	doi = {10.1007/s10115-007-0107-1},
	id = {Peng2008},
	isbn = {0219-3116},
	journal = {Knowledge and Information Systems},
	number = {3},
	pages = {281--301},
	title = {SVM based adaptive learning method for text classification from positive and unlabeled documents},
	url = {https://doi.org/10.1007/s10115-007-0107-1},
	volume = {16},
	year = {2008}}

@inproceedings{Li2003,
author = {Li, Xiaoli and Liu, Bing},
year = {2003},
month = {01},
pages = {587-594},
title = {Learning to Classify Texts Using Positive and Unlabeled Data.},
booktitle = {18th International Joint Conference on Artificial Intelligence},
publisher = {Morgan Kaufmann Publishers Inc},
address = {Acapulco, Mexico}
}

@inproceedings{Li2010,
    title = "Negative Training Data Can be Harmful to Text Classification",
    author = "Li, Xiao-Li  and
      Liu, Bing  and
      Ng, See-Kiong",
    editor = "Li, Hang  and
      M{\`a}rquez, Llu{\'\i}s",
    booktitle = "Empirical Methods in Natural Language Processing",
    year = "2010",
    url = "https://aclanthology.org/D10-1022",
    pages = "218--228",
publisher = "ACM",
address = "Cambridge Massachusetts"
}

@misc{Uden2007,
  title={Rocchio : Relevance Feedback in Learning Classification Algorithms},
  author={M. A. van Uden},
  year={2007},
  url={https://api.semanticscholar.org/CorpusID:14424311}
}

@inproceedings{Chaudhari2012,
author = {Chaudhari, Sneha and Shevade, Shirish},
title = {Learning from Positive and Unlabelled Examples Using Maximum Margin Clustering},
year = {2012},
isbn = {9783642344862},
doi = {10.1007/978-3-642-34487-9_56},
booktitle = {19th International Conference on Neural Information Processing},
pages = {465–473},
numpages = {9},
publisher = {ACM},
address = {Doha Qatar}
}

@inproceedings{Li2007,
  title={Learning to Identify Unexpected Instances in the Test Set},
  author={Xiaoli Li and B. Liu and See-Kiong Ng},
  booktitle={International Joint Conference on Artificial Intelligence},
  year={2007},
pages = {2802 - 2807},
  url={https://api.semanticscholar.org/CorpusID:14296672},
address = {Hyderabad India},
publisher = {Morgan Kaufmann Publishers Inc.}
}

@InProceedings{Li2005,
author="Li, Xiao-Li
and Liu, Bing",
title="Learning from Positive and Unlabeled Examples with Different Data Distributions",
booktitle="Machine Learning: ECML 2005",
year="2005",
pages="218--229",
isbn="978-3-540-31692-3",
address = "Porto, Portugal",
publisher = "Springer"
}

@InProceedings{Yu2007,
author="Yu, Shuang
and Li, Chunping",
title="PE-PUC: A Graph Based PU-Learning Approach for Text Classification",
booktitle="Machine Learning and Data Mining in Pattern Recognition",
year="2007",
pages="574--584",
address = "Leipzig, Germany",
publisher = "Springer"
}

@article{Ortega2023,
	author = {Ortega V{\'a}zquez, Carlos and vanden Broucke, Seppe and De Weerdt, Jochen},
	date = {2023/05/01},
	doi = {10.1007/s10618-023-00925-9},
	id = {Ortega V{\'a}zquez2023},
	isbn = {1573-756X},
	journal = {Data Mining and Knowledge Discovery},
	number = {3},
	pages = {1301--1325},
	title = {A two-step anomaly detection based method for PU classification in imbalanced data sets},
	url = {https://doi.org/10.1007/s10618-023-00925-9},
	volume = {37},
	year = {2023}
}

@inproceedings{Elkan2008,
author = {Elkan, Charles and Noto, Keith},
title = {Learning Classifiers from Only Positive and Unlabeled Data},
year = {2008},
isbn = {9781605581934},
doi = {10.1145/1401890.1401920},
booktitle = {Proceedings of the 14th ACM SIGKDD International Conference on Knowledge Discovery and Data Mining},
pages = {213–220},
numpages = {8},
publisher = {ACM},
address = {Las Vegas, Nevada, USA}
}

@inproceedings{Lee2003,
author = {Lee, Wee Sun and Liu, Bing},
title = {Learning with Positive and Unlabeled Examples Using Weighted Logistic Regression},
year = {2003},
isbn = {1577351894},
booktitle = {20th International Conference on International Conference on Machine Learning},
pages = {448–455},
numpages = {8},
publisher = {AAAI Press},
address = {Washington, DC USA}
}

@INPROCEEDINGS{Liu2003b,
  author={Liu, B. and Dai, Y. and Li, X. and Lee, W.S. and Yu, P.S.},
  booktitle={Third IEEE International Conference on Data Mining}, 
  title={Building text classifiers using positive and unlabeled examples}, 
  year={2003},
  pages={179-186},
  doi={10.1109/ICDM.2003.1250918},
publisher = {IEEE},
address = {Melbourne, Florida}
}

@InProceedings{Ke2012,
author="Ke, Ting
and Yang, Bing
and Zhen, Ling
and Tan, Junyan
and Li, Yi
and Jing, Ling",
editor="Wang, Jun
and Yen, Gary G.
and Polycarpou, Marios M.",
title="Building High-Performance Classifiers Using Positive and Unlabeled Examples for Text Classification",
booktitle="Advances in Neural Networks -- ISNN 2012",
year="2012",
pages="187--195",
isbn="978-3-642-31362-2",
publisher = "Springer",
address = "Shenyang, China"
}

@InProceedings{Liu2005,
author="Liu, Zhigang
and Shi, Wenzhong
and Li, Deren
and Qin, Qianqing",
title="Partially Supervised Classification -- Based on Weighted Unlabeled Samples Support Vector Machine",
booktitle="Advanced Data Mining and Applications",
year="2005",
pages="118--129",
isbn="978-3-540-31877-4",
address = "Wuhan, China",
publisher = "Springer"
}

@article{Mordelet2013,
  title={Supervised inference of gene regulatory networks from positive and unlabeled examples.},
  author={Fantine Mordelet and Jean-Philippe Vert},
  journal={Methods in molecular biology},
  year={2013},
  volume={939},
  pages={
          47-58
        },
  url={https://api.semanticscholar.org/CorpusID:2974814}
}

@article{Suykens1999,
	author = {Suykens, J. A. K. and Vandewalle, J.},
	date = {1999/06/01},
	doi = {10.1023/A:1018628609742},
	id = {Suykens1999},
	isbn = {1573-773X},
	journal = {Neural Processing Letters},
	number = {3},
	pages = {293--300},
	title = {Least Squares Support Vector Machine Classifiers},
	url = {https://doi.org/10.1023/A:1018628609742},
	volume = {9},
	year = {1999}}

@article{Ke2018,
	author = {Ke, Ting and Jing, Ling and Lv, Hui and Zhang, Lidong and Hu, Yaping},
	date = {2018/08/01},
	doi = {10.1007/s10489-017-1076-z},
	id = {Ke2018},
	isbn = {1573-7497},
	journal = {Applied Intelligence},
	number = {8},
	pages = {2373--2392},
	title = {Global and local learning from positive and unlabeled examples},
	url = {https://doi.org/10.1007/s10489-017-1076-z},
	volume = {48},
	year = {2018}}

@article{Ke2018b,
title = {A biased least squares support vector machine based on Mahalanobis distance for PU learning},
journal = {Physica A: Statistical Mechanics and its Applications},
volume = {509},
pages = {422-438},
year = {2018},
issn = {0378-4371},
doi = {https://doi.org/10.1016/j.physa.2018.05.128},
url = {https://www.sciencedirect.com/science/article/pii/S0378437118306794},
author = {Ting Ke and Hui Lv and Mingjing Sun and Lidong Zhang},
}

@article{Bekker2018, 
title={Estimating the Class Prior in Positive and Unlabeled Data Through Decision Tree Induction}, 
volume={32}, url={https://ojs.aaai.org/index.php/AAAI/article/view/11715}, 
DOI={10.1609/aaai.v32i1.11715}, 
number={1}, 
journal={Proceedings of the AAAI Conference on Artificial Intelligence}, 
author={Bekker, Jessa and Davis, Jesse}, 
year={2018},
pages = {2712-2719},
month={Apr.} }

@InProceedings{Ramaswamy2016,
  title = 	 {Mixture Proportion Estimation via Kernel Embeddings of Distributions},
  author = 	 {Ramaswamy, Harish and Scott, Clayton and Tewari, Ambuj},
  booktitle = 	 {Proceedings of The 33rd International Conference on Machine Learning},
  pages = 	 {2052--2060},
  year = 	 {2016},
  volume = 	 {48},
  series = 	 {Proceedings of Machine Learning Research},
  address = 	 {New York, New York, USA},
  url = 	 {https://proceedings.mlr.press/v48/ramaswamy16.html},
publisher = {JMLR}
  }

@article{Plessis2017,
	author = {du Plessis, Marthinus C. and Niu, Gang and Sugiyama, Masashi},
	date = {2017/04/01},
	doi = {10.1007/s10994-016-5604-6},
	id = {du Plessis2017},
	isbn = {1573-0565},
	journal = {Machine Learning},
	number = {4},
	pages = {463--492},
	title = {Class-prior estimation for learning from positive and unlabeled data},
	url = {https://doi.org/10.1007/s10994-016-5604-6},
	volume = {106},
	year = {2017}}

@article{Zeiberg2020, 
title={Fast Nonparametric Estimation of Class Proportions in the Positive-Unlabeled Classification Setting}, volume={34}, 
url={https://ojs.aaai.org/index.php/AAAI/article/view/6151}, 
DOI={10.1609/aaai.v34i04.6151}, 
number={04}, 
journal={Proceedings of the AAAI Conference on Artificial Intelligence}, 
author={Zeiberg, Daniel and Jain, Shantanu and Radivojac, Predrag}, 
year={2020}, 
month={Apr.}, 
pages={6729-6736} }

@inproceedings{Plessis2014,
 author = {du Plessis, Marthinus C and Niu, Gang and Sugiyama, Masashi},
 booktitle = {Advances in Neural Information Processing Systems},
 title = {Analysis of Learning from Positive and Unlabeled Data},
 volume = {27},
 year = {2014},
publisher = {Curran Associates, Inc},
address = {Montreal, Canada},
pages = {9}
}

@InProceedings{Plessis2015,
  title = 	 {Convex Formulation for Learning from Positive and Unlabeled Data},
  author = 	 {Plessis, Marthinus Du and Niu, Gang and Sugiyama, Masashi},
  booktitle = 	 {32nd International Conference on Machine Learning},
  pages = 	 {1386--1394},
  year = 	 {2015},
  volume = 	 {37},
  series = 	 {Proceedings of Machine Learning Research},
address = {Lille France},
publisher = {JMLR}
  }

@InProceedings{Kiryo2017,
author = {Kiryo, Ryuichi and Niu, Gang and Plessis, Marthinus and Sugiyama, Masashi},
year = {2017},
month = {03},
pages = {},
title = {Positive-Unlabeled Learning with Non-Negative Risk Estimator},
address = {Long Beach, CA, USA},
publisher = {Curran Associates Inc.},
booktitle = {Advances in Neural Information Processing Systems}
}

@inproceedings{Su2021,
  title     = {Positive-Unlabeled Learning from Imbalanced Data},
  author    = {Su, Guangxin and Chen, Weitong and Xu, Miao},
  booktitle = {13th International Joint Conference on
               Artificial Intelligence, {IJCAI-21}},
  pages     = {2995--3001},
  year      = {2021},
  month     = {8},
  doi       = {10.24963/ijcai.2021/412},
address = {Virtual conference},
publisher = {Curran Associates, Inc}
}

@inproceedings{Niu2016,
    author = {Niu, G. and du Plessis, M.c. and Sakai, T. and Ma, Y. and Sugiyama. M.},
    title = {Theoretical comparisons of positive-unlabeled learning against positive-negative learning},
    booktitle = {NIPS},
    year = {2016},
address = {Barcelona Spain},
publisher = {Curran Associates Inc.},
pages = {9}
}

@ARTICLE{Lin2020,
  author={Lin, Tsung-Yi and Goyal, Priya and Girshick, Ross and He, Kaiming and Dollár, Piotr},
  journal={IEEE Transactions on Pattern Analysis and Machine Intelligence}, 
  title={Focal Loss for Dense Object Detection}, 
  year={2020},
  volume={42},
  number={2},
  pages={318-327},
  doi={10.1109/TPAMI.2018.2858826}}

@article{Cecchini2010,
  title={Detecting management fraud in public companies},
  author={Cecchini, Mark and Aytug, Haldun and Koehler, Gary J and Pathak, Praveen},
  journal={Management Science},
  volume={56},
  number={7},
  pages={1146--1160},
  year={2010},
  publisher={INFORMS}
}

@article{Dechow2011,
  title={Predicting material accounting misstatements},
  author={Dechow, Patricia M and Ge, Weili and Larson, Chad R and Sloan, Richard G},
  journal={Contemporary accounting research},
  volume={28},
  number={1},
  pages={17--82},
  year={2011},
  publisher={Wiley Online Library}
}

@article{Bao2020,
  title={Detecting accounting fraud in publicly traded US firms using a machine learning approach},
  author={Bao, Yang and Ke, Bin and Li, Bin and Yu, Y Julia and Zhang, Jie},
  journal={Journal of Accounting Research},
  volume={58},
  number={1},
  pages={199--235},
  year={2020},
  publisher={Wiley Online Library}
}

@article{Bertomeu2021,
 title={Using machine learning to detect misstatements},
 author={Bertomeu, J. and Cheynel, E. and Floyd, E and Pan W.},
 journal={Review of Accounting Studies},
 volume={26},
 pages={468–519},
 year={2021}
}

@article{izavitsESWA,
    title = {Calibrating TabTransformer for financial misstatement detection},
    author = {Zavitsanos, Elias and Kelesis, Dimitrios and Paliouras, Georgios},
    journal = {Applied Intelligence},
    year = {2025},
    volume = {55},
    number = {1},
pages = {15}
    
}

@article{Dyck2010,
  title={Who blows the whistle on corporate fraud?},
  author={Dyck, Alexander and Morse, Adair and Zingales, Luigi},
  journal={The journal of finance},
  volume={65},
  number={6},
  pages={2213--2253},
  year={2010},
  publisher={Wiley Online Library}
}

@inproceedings{gmlp2021,
 author = {Liu, Hanxiao and Dai, Zihang and So, David and Le, Quoc V},
 booktitle = {Advances in Neural Information Processing Systems},
 pages = {9204--9215},
 title = {Pay Attention to MLPs},
 volume = {34},
 year = {2021},
publisher = {Curran Associates, Inc.},
address = {Virtual conference}
 }

@misc{tabtransformer2020,
  doi = {10.48550/ARXIV.2012.06678},
  author = {Huang, Xin and Khetan, Ashish and Cvitkovic, Milan and Karnin, Zohar},
  title = {TabTransformer: Tabular Data Modeling Using Contextual Embeddings},
  publisher = {arXiv},
  year = {2020},
  copyright = {Creative Commons Zero v1.0 Universal}
}

@inproceedings{Chou2020,
author = {Chou, Yu-Ting and Niu, Gang and Lin, Hsuan-Tien and Sugiyama, Masashi},
title = {Unbiased risk estimators can mislead: a case study of learning with complementary labels},
year = {2020},
booktitle = {Proceedings of the 37th International Conference on Machine Learning},
publisher = {JMLR},
address = {Virtual conference},
pages = {10}
}

@inproceedings{FocalLossTheory2020,
author = {Charoenphakdee, Nontawat and Vongkulbhisal, Jayakorn and Chairatanakul, Nuttapong and Sugiyama, Masashi},
booktitle = {IEEE Conference on Computer Vision and Pattern Recognition},
year = {2020},
month = {11},
pages = {},
title = {On Focal Loss for Class-Posterior Probability Estimation: A Theoretical Perspective},
publisher = {IEEE},
address = {Nashville, TN, USA}
}

@article{Saerens2002,
    author = {Saerens, Marco and Latinne, Patrice and Decaestecker, Christine},
    title = "{Adjusting the Outputs of a Classifier to New a Priori Probabilities: A Simple Procedure}",
    journal = {Neural Computation},
    volume = {14},
    number = {1},
    pages = {21-41},
    year = {2002},
    month = {01},
    issn = {0899-7667}
}

@article{Plessis2014b,
title = {Semi-supervised learning of class balance under class-prior change by distribution matching},
journal = {Neural Networks},
volume = {50},
pages = {110-119},
year = {2014},
issn = {0893-6080},
doi = {https://doi.org/10.1016/j.neunet.2013.11.010},
url = {https://www.sciencedirect.com/science/article/pii/S0893608013002748},
author = {Marthinus Christoffel {du Plessis} and Masashi Sugiyama}
}

@INPROCEEDINGS{Yoo2021,
  author={Yoo, Jaemin and Kim, Junghun and Yoon, Hoyoung and Kim, Geonsoo and Jang, Changwon and Kang, U},
  booktitle={2021 IEEE International Conference on Data Mining (ICDM)}, 
  title={Accurate Graph-Based PU Learning without Class Prior}, 
  year={2021},
  volume={},
  number={},
  pages={827-836},
doi = {10.1109/ICDM51629.2021.00094},
address = {Virtual conference},
publisher = {IEEE}
}

@article{PUHellinger,
	author = {Ortega V{\'a}zquez, Carlos and vanden Broucke, Seppe and De Weerdt, Jochen},
	doi = {10.1007/s10994-023-06323-y},
	journal = {Machine Learning},
	number = {7},
	pages = {4547--4578},
	title = {Hellinger distance decision trees for PU learning in imbalanced data sets},
	volume = {113},
	year = {2024},
	}

@ARTICLE{LBE2022,
author={Gong, Chen and Wang, Qizhou and Liu, Tongliang and Han, Bo and You, Jane and Yang, Jian and Tao, Dacheng},
journal={ IEEE Transactions on Pattern Analysis \& Machine Intelligence },
title={{ Instance-Dependent Positive and Unlabeled Learning With Labeling Bias Estimation }},
year={2022},
volume={44},
number={08},
ISSN={1939-3539},
pages={4163-4177},
doi={10.1109/TPAMI.2021.3061456},
url = {https://doi.ieeecomputersociety.org/10.1109/TPAMI.2021.3061456},
publisher={IEEE Computer Society},
address={Los Alamitos, CA, USA},
month=aug}

@misc{foster2019ellinftyvectorcontractionrademacher,
      title={$\ell_{\infty}$ Vector Contraction for Rademacher Complexity}, 
      author={Dylan J. Foster and Alexander Rakhlin},
      year={2019},
      eprint={1911.06468},
      archivePrefix={arXiv},
      primaryClass={cs.LG},
      url={https://arxiv.org/abs/1911.06468}, 
}

@article{Bartlett2003,
author = {Bartlett, Peter L. and Mendelson, Shahar},
title = {Rademacher and gaussian complexities: risk bounds and structural results},
year = {2003},
issue_date = {3/1/2003},
publisher = {JMLR.org},
volume = {3},
number = {null},
issn = {1532-4435},
journal = {J. Mach. Learn. Res.},
month = mar,
pages = {463–482},
numpages = {20}
}

@inproceedings{izavits2025,
author = {Zavitsanos, Elias and Bougiatiotis, Konstantinos and Sideras, Andreas and Paliouras, Georgios},
title = {Positive-Unlabeled Learning for Financial Misstatement Detection under Realistic Constraints},
year = {2025},
isbn = {9798400722202},
publisher = {ACM},
address = {New York, NY, USA},
doi = {10.1145/3768292.3770366},
booktitle = {Proceedings of the 6th ACM International Conference on AI in Finance},
pages = {864–872},
numpages = {9},
location = {
},
series = {ICAIF '25}
}

@article{hollmann2025tabpfn,
 title={Accurate predictions on small data with a tabular foundation model},
 author={Hollmann, Noah and M{\"u}ller, Samuel and Purucker, Lennart and
         Krishnakumar, Arjun and K{\"o}rfer, Max and Hoo, Shi Bin and
         Schirrmeister, Robin Tibor and Hutter, Frank},
 journal={Nature},
 year={2025},
 month={01},
 day={09},
 doi={10.1038/s41586-024-08328-6},
 publisher={Springer Nature},
 url={https://www.nature.com/articles/s41586-024-08328-6},
}

\end{document}